# Flood Prediction Using Machine Learning Models: Literature Review

**Amir Mosavi [1,*], Pinar Ozturk [1] and Kwok-wing Chau [2]**

[1] Department of Computer Science (IDI), Norwegian University of Science and Technology (NTNU), Trondheim, NO-7491, Norway

[2] Department of Civil and Environmental Engineering, Hong Kong Polytechnic University, Hong Kong, China; dr.kwok-wing.chau@polyu.edu.hk

*Correspondence: amir.mosavi@ntnu.no

**Abstract:** Floods are among the most destructive natural disasters, which are highly complex to model. The research on the advancement of flood prediction models contributed to risk reduction, policy suggestion, minimization of the loss of human life, and reduction the property damage associated with floods. To mimic the complex mathematical expressions of physical processes of floods, during the past two decades, machine learning (ML) methods contributed highly in the advancement of prediction systems providing better performance and cost-effective solutions. Due to the vast benefits and potential of ML, its popularity dramatically increased among hydrologists. Researchers through introducing novel ML methods and hybridizing of the existing ones aim at discovering more accurate and efficient prediction models. The main contribution of this paper is to demonstrate the state of the art of ML models in flood prediction and to give insight into the most suitable models. In this paper, the literature where ML models were benchmarked through a qualitative analysis of robustness, accuracy, effectiveness, and speed are particularly investigated to provide an extensive overview on the various ML algorithms used in the field. The performance comparison of ML models presents an in-depth understanding of the different techniques within the framework of a comprehensive evaluation and discussion. As a result, this paper introduces the most promising prediction methods for both long-term and short-term floods. Furthermore, the major trends in improving the quality of the flood prediction models are investigated. Among them, hybridization, data decomposition, algorithm ensemble, and model optimization are reported as the most effective strategies for the improvement of ML methods. This survey can be used as a guideline for hydrologists as well as climate scientists in choosing the proper ML method according to the prediction task.

**Keywords:** flood prediction; flood forecasting; flash-flood model, big flood management; hydrologic model; rainfall–runoff, hybrid & ensemble machine learning; artificial neural networks (ANNs); support vector machines (SVM); natural hazards & disasters; adaptive neuro-fuzzy inference system (ANFIS); decision trees (DT); internet of things (IoT); random forest (RF); survey; classification and regression trees (CART), data science; deep learning; big data; bagging, boosting, artificial intelligence (AI); soft computing; extreme event management; time series prediction; multilayer perceptron (MLP); simulated annealing (SA); multivariate adaptive regression splines (MARS), supervised learning

## 1. Introduction

Among the natural disasters, floods are the most destructive, causing massive damage to human life, infrastructure, agriculture, and the socioeconomic system. Governments, therefore, are under pressure to develop reliable and accurate maps of flood risk areas and further plan for sustainable flood risk management focusing on prevention, protection, and preparedness [1]. Flood prediction models are of significant importance for hazard assessment and extreme event management. Robust and accurate prediction contribute highly to water recourse management strategies, policy suggestions and analysis, and further evacuation modeling [2]. Thus, the importance of advanced systems for short-term and long-term prediction for flood and other hydrological events is strongly emphasized to alleviate damage [3]. However, the prediction of flood lead time and occurrence location is fundamentally complex due to the dynamic nature of climate condition. Therefore, today's major flood prediction models are mainly data-specific and involve various simplified assumptions [4]. Thus, to mimic the complex mathematical expressions of physical processes and basin behavior, such models benefit from specific techniques e.g., event-driven, empirical black box, lumped and distributed, stochastic, deterministic, continuous, and hybrids [5].

Physically based models [6] were long used to predict hydrological events, such as storm [7,8], rainfall/runoff [9,10], shallow water condition [11], hydraulic models of flow [12,13], and further global circulation phenomena [14], including the coupled effects of atmosphere, ocean, and floods [15]. Although physical models showed great capabilities for predicting a diverse range of flooding scenarios, they often require various types of hydro-geomorphological monitoring datasets, requiring intensive computation, which prohibits short-term prediction [16]. Furthermore, as stated in Reference [17], the development of physically based models often requires in-depth knowledge and expertise regarding hydrological parameters, reported to be highly challenging. Moreover, numerous studies suggest that there is a gap in short-term prediction capability of physical models (Costabile and Macchione [15]). For instance, on many occasions, such models failed to predict properly [18]. Van den Honert and McAneney [18] documented the failure in the prediction of floods accrued in Queensland, Australia in 2010. Similarly, numerical prediction models [19] were reported in the advancement of deterministic calculations, and were not reliable due to systematic errors [20]. Nevertheless, major improvements in physically based models of flood were recently reported through the hybridization of models [21], as well as advanced flow simulations [22,23].

In addition to numerical and physical models, data-driven models also have a long tradition in flood modeling, which recently gained more popularity. Data-driven methods of prediction assimilate the measured climate indices and hydro-meteorological parameters to provide better insight. Among them, statistical models of autoregressive moving average (ARMA) [24], multiple linear regression (MLR) [25], and autoregressive integrated moving average (ARIMA) [26] are the most common flood frequency analysis (FFA) methods for modeling flood prediction. FFA was among the early statistical methods for predicting floods [27]. Regional flood frequency analyses (RFFA) [28], more advanced versions, were reported to be more efficient when compared to physical models considering computation cost and generalization. Assuming floods as stochastic processes, they can be predicted using certain probability distributions from historical streamflow data [29]. For instance, the climatology average method (CLIM) [28], empirical orthogonal function (EOF) [30], multiple linear regressions (MLR), quantile regression techniques (QRT) [31], and Bayesian forecasting models [32] are widely used for predicting major floods. However, they were reported to be unsuitable for short-term prediction, and, in this context, they need major improvement due to the lack of accuracy, complexity of the usage, computation cost, and robustness of the method. Furthermore, for reliable long-term prediction, at least, a decade of data from measurement gauges should be analyzed for a meaningful forecast [32]. In the absence of such a dataset, however, FFA can be done using hydrologic models of RFFA, e.g., MISBA [33] and Sacramento [34], as reliable empirical methods with regional applications, where streamflow measurements are unavailable. In this context, distributed numerical models are used as an attractive solution [35]. Nonetheless, they do not provide quantitative flood predictions, and their forecast skill level is “only moderate” and they lack accuracy [36].

The drawbacks of the physically based and statistical models mentioned above encourage the usage of advanced data-driven models, e.g., machine learning (ML). A further reason for the popularity of such models is that they can numerically formulate the flood nonlinearity, solely based on historical data without requiring knowledge about the underlying physical processes. Data-driven prediction models using ML are promising tools as they are quicker to develop with minimal inputs. ML is a field of artificial intelligence (AI) used to induce regularities and patterns, providing easier implementation with low computation cost, as well as fast training, validation, testing, and evaluation, with high performance compared to physical models, and relatively less complexity [37]. The continuous advancement of ML methods over the last two decades demonstrated their suitability for flood forecasting with an acceptable rate of outperforming conventional approaches [38]. A recent investigation by Reference [39], which compared performance of a number of physical and ML prediction models, showed a higher accuracy of ML models. Furthermore, the literature includes numerous successful experiments of quantitative precipitation forecasting (QPF) using ML methods for different lead-time predictions [40,41]. In comparison to traditional statistical models, ML models were used for prediction with greater accuracy [42]. Ortiz-García et al. [43] described how ML techniques could efficiently model complex hydrological systems such as floods. Many ML algorithms, e.g., artificial neural networks (ANNs) [44], neuro-fuzzy [45,46], support vector machine (SVM) [47], and support vector regression (SVR) [48,49], were reported as effective for both short-term and long-term flood forecast. In addition, it was shown that the performance of ML could be improved through hybridization with other ML methods, soft computing techniques, numerical simulations, and/or physical models. Such applications provided more robust and efficient models that can effectively learn complex flood systems in an adaptive manner. Although the literature includes numerous evaluation performance analyses of individual ML models [49–52], there is no definite conclusion reported with regards to which models function better in certain applications. In fact, the literature includes only a limited number of surveys on specific ML methods in specific hydrology fields [53–55]. Consequently, there is a research gap for a comprehensive literature review in the general applications of ML in all flood resource variables from the perspective of ML modeling and data-driven prediction systems.

Nonetheless, ML algorithms have important characteristics that need to be carefully taken into consideration. The first is that they are as good as their training, whereby the system learns the target task based on past data. If the data is scarce or does not cover varieties of the task, their learning falls short, and hence, they cannot perform well when they are put into work. Therefore, using robust data enrichment is essential through, e.g., implementing a distribution function of sums of weights [56], invariance assessments to retain the group characteristics [57], or recovering the missing variables using causally dependent coefficients [58].

The second aspect is the capability of each ML algorithm, which may vary across different types of tasks. This can also be called a "generalization problem", which indicates how well the trained system can predict cases it was not trained for, i.e., whether it can predict beyond the range of the training dataset. For example, some algorithms may perform well for short-term predictions, but not for long-term predictions. These characteristics of the algorithms need to be clarified with respect to the type and amount of available training data, and the type of prediction task, e.g., water level and streamflow. In this review, we look into examples of the use of various ML algorithms for various types of tasks. At the abstract level, we decided to divide the target tasks into short-term and long-term prediction. We then reviewed ML applications for flood-related tasks, where we structured ML methods as single methods and hybrid methods. Hybrid methods are those that combine more than one ML method.

Here, we should note that this paper surveys ML models used for predictions of floods on sites where rain gauges or intelligent sensing systems used. Our goal was to survey prediction models with various lead times to floods at a particular site. From this perspective, spatial flood prediction was not involved in this study, as we did not study prediction models used to estimate/identify the location of floods. In fact, we were concerned only with the lead time for an identified site.

## 2. Method and Outline

This survey identifies the state of the art of ML methods for flood prediction where peer-reviewed articles in top-level subject fields are reviewed. Among the articles identified, through search queries using the search strategy, those including the performance evaluation and comparison of ML methods were given priority to be included in the review to identify the ML methods that perform better in particular applications. Furthermore, to choose an article, four types of quality measure for each article were considered, i.e., source normalized impact per paper (SNIP), CiteScore, SCImago journal rank (SJR), and h-index. The papers were reviewed in terms of flood resource variables, ML methods, prediction type, and the obtained results.

The applications in flood prediction can be classified according to flood resource variables, i.e., water level, river flood, soil moisture, rainfall–discharge, precipitation, river inflow, peak flow, river flow, rainfall–runoff, flash flood, rainfall, streamflow, seasonal stream flow, flood peak discharge, urban flood, plain flood, groundwater level, rainfall stage, flood frequency analysis, flood quantiles, surge level, extreme flow, storm surge, typhoon rainfall, and daily flows [59]. Among these key influencing flood resource variables, rainfall and the spatial examination of the hydrologic cycle had the most remarkable role in runoff and flood modeling [60]. This is the reason why quantitative rainfall prediction, including avalanches, slush flow, and melting snow, is traditionally used for flood prediction, especially in the prediction of flash floods or short-term flood prediction [61]. However, rainfall prediction was shown to be inadequate for accurate flood prediction. For instance, the prediction of streamflow in a long-term flood prediction scenario depends on soil moisture estimates in a catchment, in addition to rainfall [62]. Although, high-resolution precipitation forecasting is essential, other flood resource variables were considered in the [63]. Thus, the methodology of this literature review aims to include the most effective flood resource variables in the search queries.

A combination of these flood resource variables and ML methods was used to implement the complete list of search queries. Note that the ML methods for flood prediction may vary significantly according to the application, dataset, and prediction type. For instance, ML methods used for short-term water level prediction are significantly different from those used for long-term streamflow prediction. Figure 1 represents the organization of the search queries and further describes the survey search methodology.

The search query included three main search terms. The flood resource variables were considered as term 1 of the search (<Flood resource variable1-n>), which included 25 keywords for search queries mentioned above. Term 2 of search (<ML method1-m>) included the ML algorithms. The collection of the references [16,26,28,37,38,42,44] provides a complete list of ML methods, from which the 25 most popular algorithms in engineering applications were used as the keywords of this search. Term 3 included the four search terms most often used in describing flood prediction, i.e., "prediction", "estimation", "forecast", or "analysis". The total search resulted in 6596 articles. Among them, 180 original research papers were refined through our quality measure included in the survey.

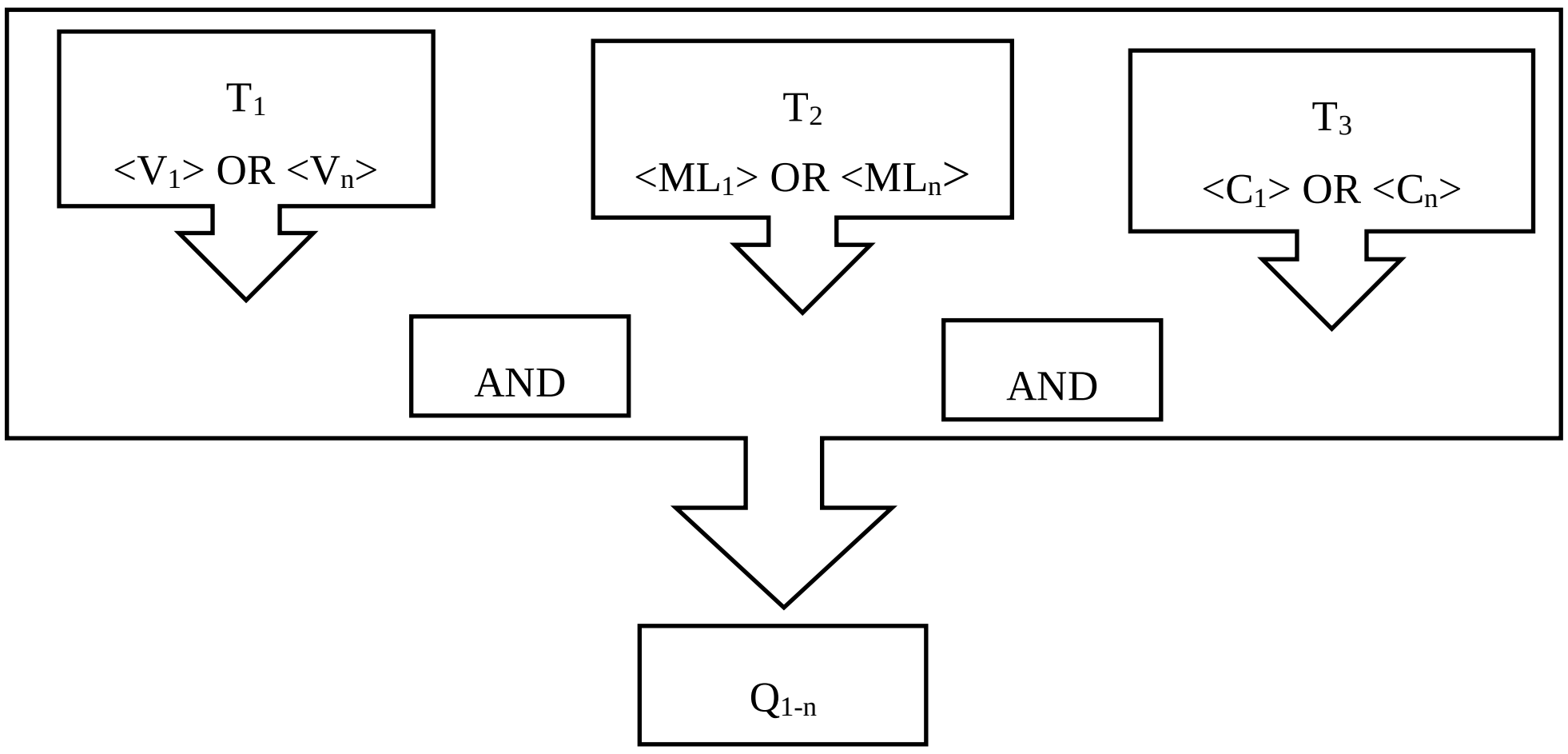

**Figure 1.** Flowchart of the search queries.

Section 3 presents the state of the art of ML in flood prediction. A technical description on the ML method and a brief background in flood applications are provided. Section 4 presents the survey of ML methods used for short-term flood prediction. Section 5 presents the survey of ML methods used for long-term flood prediction. Section 6 presents the conclusions.

## 3. State of the Art of ML Methods in Flood Prediction

For creating the ML prediction model, the historical records of flood events, in addition to real-time cumulative data of a number of rain gauges or other sensing devices for various return periods, are often used. The sources of the dataset are traditionally rainfall and water level, measured either by ground rain gauges, or relatively new remote-sensing technologies such as satellites, multisensor systems, and/or radars [62]. Nevertheless, remote sensing is an attractive tool for capturing higher-resolution data in real time. In addition, the high resolution of weather radar observations often provides a more reliable dataset compared to rain gauges [63]. Thus, building a prediction model based on a radar rainfall dataset was reported to provide higher accuracy in general [64]. Whether using a radar-based dataset or ground gauges to create a prediction model, the historical dataset of hourly, daily, and/or monthly values is divided into individual sets to construct and evaluate the learning models. To do so, the individual sets of data undergo training, validation, verification, and testing. The principle behind the ML modeling workflow and the strategy for flood modeling are described in detail in the literature [48,65]. Figure 2 represents the basic flow for building an ML model. The major ML algorithms applied to flood prediction include ANNs [66], neuro-fuzzy [67], adaptive neuro-fuzzy inference systems (ANFIS) [68], support vector machines (SVM) [69], wavelet neural networks (WNN) [70], and multilayer perceptron (MLP) [71]. In the following subsections, a brief description and background of these fundamental ML algorithms are presented.

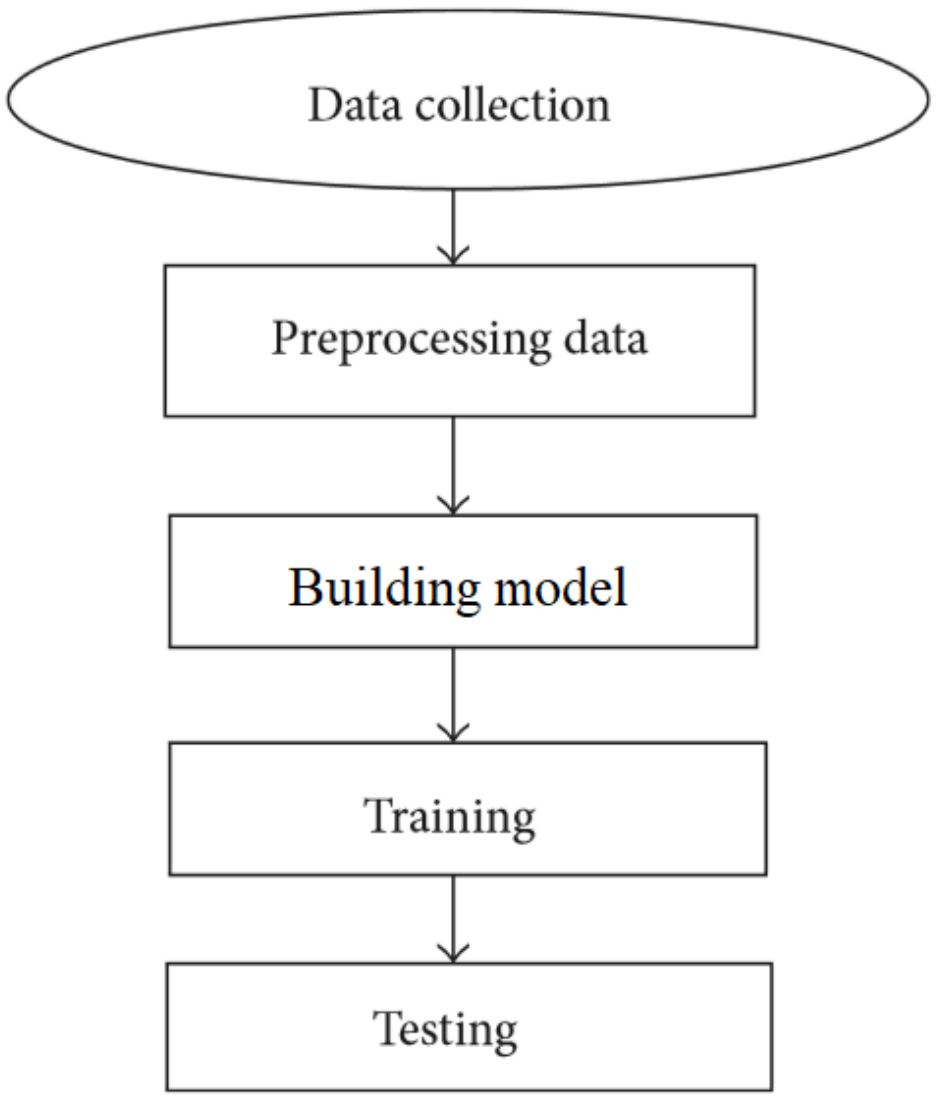

**Figure 2.** Basic flow for building the machine learning (ML) model.

### *3.1. Artificial Neural Networks (ANNs)*

ANNs are efficient mathematical modeling systems with efficient parallel processing, enabling them to mimic the biological neural network using inter-connected neuron units. Among all ML methods, ANNs are the most popular learning algorithms, known to be versatile and efficient in modeling complex flood processes with a high fault tolerance and accurate approximation [39]. In comparison to traditional statistical models, the ANN approach was used for prediction with greater accuracy [72]. ANN algorithms are the most popular for modeling flood prediction since their first usage in the 1990s [73]. Instead of a catchment's physical characteristics, ANNs derive meaning from historical data. Thus, ANNs are considered as reliable data-driven tools for constructing black-box models of complex and nonlinear relationships of rainfall and flood [74], as well as river flow and discharge forecasting [75]. Furthermore, a number of surveys (e.g., Reference [76]) suggest ANN as one of the most suitable modeling techniques which provide an acceptable generalization ability and speed compared to most conventional models. References [77,78] provided reviews on ANN applications in flood. ANNs were already successfully used for numerous flood prediction applications, e.g., streamflow forecasting [79], river flow [80,81], rainfall–runoff [82], precipitation–runoff modeling [83], water quality [55], evaporation [56], river stage prediction [84], low-flow estimation [85], river flows [86], and river time series [57]. Despite the advantages of ANNs, there are a number drawbacks associated with using ANNs in flood modeling, e.g., network architecture, data handling, and physical interpretation of the modeled system. A major drawback when using ANNs is the relatively low accuracy, the urge to iterate parameter tuning, and the slow response to gradient-based learning processes [87]. Further drawbacks associated with ANNs include precipitation prediction [88,89] and peak-value prediction [90].

The feed-forward neural network (FFNN) [25] is a class of ANN, whereby the network's connections are not in cyclical form. FFNNs are the simplest type of ANN, whereby information moves in a forward direction from input nodes to the hidden layer and later to output nodes. On the other hand, a recurrent neural network (RNN) [91] is a class of ANN, whereby the network's connections form a time sequence for dynamic temporal behavior. Furthermore, RNNs benefit from extra memory to analyze input sequences. In ANNs, backpropagation (BP) is a multi-layered NN where weights are calculated using the propagation of the backward error gradient. In BP, there are more phases in the learning cycle, using a function for activation to send signals to the other nodes. Among various ANNs, the backpropagation ANN (BPNN) was identified as the most powerful prediction tool suitable for flood time-series prediction [26]. Extreme learning machine (ELM) [92] is an easy-to-use form of FFNN, with a single hidden layer. Here, ELM was studied under the scope of ANN methods. ELM for flood prediction recently became of interest for hydrologists and was used to model short-term streamflow with promising results [93,94].

### *3.2. Multilayer Perceptron (MLP)*

The vast majority of ANN models for flood prediction are often trained with a BPNN [95]. While BPNNs are today widely used in this realm, the MLP—an advanced representation of ANNs— recently gained popularity [96]. The MLP [97] is a class of FFNN which utilizes the supervised learning of BP for training the network of interconnected nodes of multiple layers. Simplicity, nonlinear activation, and a high number of layers are characteristics of the MLP. Due to these characteristics, the model was widely used in flood prediction and other complex hydrogeological models [98]. In an assessment of ANN classes used in flood modeling, MLP models were reported to be more efficient with better generalization ability. Nevertheless, the MLP is generally found to be more difficult to optimize [99]. Back-percolation learning algorithms are used to individually calculate the propagation error in hidden network nodes for a more advanced modeling approach.

Here, it is worth mentioning that the MLP, more than any other variation of ANNs (e.g., FFNN, BPNN, and FNN), gained popularity among hydrologists. Furthermore, due to the vast number of case studies using the standard form of MLP, it diverged from regular ANNs. In addition, the authors of articles in the realm of flood prediction using the MLP refer to their models as MLP models. From this perspective, we decided to devote a separate section to the MLP.

### *3.3. Adaptive Neuro-Fuzzy Inference System (ANFIS)*

The fuzzy logic of Zadeh [100] is a qualitative modeling scheme with a soft computing technique using natural language. Fuzzy logic is a simplified mathematical model, which works on incorporating expert knowledge into a fuzzy inference system (FIS). An FIS further mimics human learning through an approximation function with less complexity, which provides great potential for nonlinear modeling of extreme hydrological events [101,102], particularly floods [103]. For instance, Reference [104] studied river level forecasting using an FIS, as did Lohani et al. (2011) [4] for rainfall–runoff modeling for water level. As an advanced form of fuzzy-rule-based modeling, neuro-fuzzy presents a hybrid of the BPNN and the widely used least-square error method [46]. The Takagi–Sugeno (T–S) fuzzy modeling technique [4], which is created using neuro-fuzzy clustering, is also widely applied in RFFA [28].

Adaptive neuro-FIS, or so-called ANFIS, is a more advanced form of neuro-fuzzy based on the T–S FIS, first coined [67,77]. Today, ANFIS is known to be one of the most reliable estimators for complex systems. ANFIS technology, through combining ANN and fuzzy logic, provides higher capability for learning [101]. This hybrid ML method corresponds to a set of advanced fuzzy rules suitable for modeling flood nonlinear functions. An ANFIS works by applying neural learning rules for identifying and tuning the parameters and structure of an FIS. Through ANN training, the ANFIS aims at catching the missing fuzzy rules using the dataset [67]. Due to fast and easy implementation, accurate learning, and strong generalization abilities, ANFIS became very popular in flood modeling. The study of Lafdani et al. [60] further described its capability in modeling short-term rainfall forecasts with high accuracy, using various types of streamflow, rainfall, and precipitation data. Furthermore, the results of Shu and [67] showed easier implementation and better generalization capability, using the one-pass subtractive clustering algorithm, which led several rounds of random selection being avoided.

### *3.4. Wavelet Neural Network (WNN)*

Wavelet transform (WT) [46] is a mathematical tool which can be used to extract information from various data sources by analyzing local variations in time series [50]. In fact, WT has significantly positive effects on modeling performance [105]. Wavelet transforms supports the reliable decomposition of an original time series to improve data quality. The accuracy of prediction is improved through discrete WT (DWT), which decomposes the original data into bands, leading to an improvement of flood prediction lead times [106]. DWT decomposes the initial data set into individual resolution levels for extracting better-quality data for model building. DWTs, due to their beneficial characteristics, are widely used in flood time-series prediction. In flood modeling, DWTs were widely applied in, e.g., rainfall–runoff [51[, daily streamflow [106], and reservoir inflow [107]. Furthermore, hybrid models of DWTs, e.g., wavelet-based neural networks (WNNs) [108], which combine WT and FFNNs, and wavelet-based regression models [109], which integrate WT and multiple linear regression (MLR), were used in time-series predictions of floods [110]. The application of WNN for flood prediction was reviewed in Reference [70], where it was concluded that WNNs can highly enhance model accuracy. In fact, most recently, WNNs, due to their potential in enhancing time-series data, gained popularity in flood modeling [50], for applications such as daily flow [111], rainfall–runoff [112], water level [113], and flash floods [114].

### *3.5. Support Vector Machine (SVM)*

Hearst et al. [115] proposed and classified the support vector (SV) as a nonlinear search algorithm using statistical learning theory. Later, the SVM [116] was introduced as a class of SV, used to minimize over-fitting and reduce the expected error of learning machines. SVM is greatly popular in flood modeling; it is a supervised learning machine which works based on the statistical learning theory and the structural risk minimization rule. The training algorithm of SVM builds models that assign new non-probabilistic binary linear classifiers, which minimize the empirical classification error and maximize the geometric margin via inverse problem solving. SVM is used to predict a quantity forward in time based on training from past data. Over the past two decades, the SVM was also extended as a regression tool, known as support vector regression (SVR) [117].

SVMs are today know as robust and efficient ML algorithms for flood prediction [118]. SVM and SVR emerged as alternative ML methods to ANNs, with high popularity among hydrologists for flood prediction. They use the statistical learning theory of structural risk minimization (SRM), which provides a unique architecture for delivering great generalization and superior efficiency. Most importantly, SVMs are both suitable for linear and nonlinear classification, and the efficient mapping of inputs into feature spaces [119]. Thus, they were applied in numerous flood prediction cases with promising results, excellent generalization ability, and better performance, compared to ANNs and MLRs, e.g., extreme rainfall [120], precipitation [43], rainfall–runoff [121], reservoir inflow [122], streamflow [123], flood quantiles [48], flood time series [124], and soil moisture [125]. Unlike ANNs, SVMs are more suitable for nonlinear regression problems, to identify the global optimal solution in flood models [126]. Although the high computation cost of using SVMs and their unrealistic outputs might be demanding, due to their heuristic and semi-black-box nature, the least-square support vector machine (LS-SVM) highly improved performance with acceptable computational efficiency [127]. The alternative approach of LS-SVM involves solving a set of linear tasks instead of complex quadratic problems [128]. Nevertheless, there are still a number of drawbacks that exist, especially in the application of seasonal flow prediction using LS-SVM [129].

### *3.6. Decision Tree (DT)*

The ML method of DT is one of the contributors in predictive modeling with a wide application in flood simulation. DT uses a tree of decisions from branches to the target values of leaves. In classification trees (CT), the final variables in a DT contain a discrete set of values where leaves represent class labels and branches represent conjunctions of features labels. When the target variable in a DT has continuous values and an ensemble of trees is involved, it is called a regression tree (RT) [130]. Regression and classification trees share some similarities and differences. As DTs are classified as fast algorithms, they became very popular in ensemble forms to model and predict floods [131]. The classification and regression tree (CART) [132,133], which is a popular type of DT used in ML, was successfully applied to flood modeling; however, its applicability to flood prediction is yet to be fully investigated [134]. The random forests (RF) method [69,135] is another popular DT method for flood prediction [136]. RF includes a number of tree predictors. Each individual tree creates a set of response predictor values associated with a set of independent values. Furthermore, an ensemble of these trees selects the best choice of classes [69]. Reference [137] introduced RF as an effective alternative to SVM, which often delivers higher performance in flood prediction modeling. Later, Bui et al. [138] compared the performances of ANN, SVM, and RF in general applications to floods, whereby RF delivered the best performance. Another major DT is the M5 decision-tree algorithm [139]. M5 constructs a DT by splitting the decision space and single attributes, thereby decreasing the variance of the final variable. Further DT algorithms popular in flood prediction include reduced-error pruning trees (REPTs), Naïve Bayes trees (NBTs), chi-squared automatic interaction detectors (CHAIDs), logistic model trees (LMTs), alternating decision trees (ADTs), and exhaustive CHAIDs (E-CHAIDs).

### *3.7. Ensemble Prediction Systems (EPSs)*

A multitude of ML modeling options were introduced for flood modeling with a strong background [140]. Thus, there is an emerging strategy to shift from a single model of prediction to an ensemble of models suitable for a specific application, cost, and dataset. ML ensembles consist of a finite set of alternative models, which typically allow more flexibility than the alternatives. Ensemble ML methods have a long tradition in flood prediction. In recent years, ensemble prediction systems (EPSs) [141] were proposed as efficient prediction systems to provide an ensemble of $N$ forecasts. In EPS, $N$ is the number of independent realizations of a model probability distribution. EPS models generally use multiple ML algorithms to provide higher performance using an automated assessment and weighting system [140]. Such a weighting procedure is carried out to accelerate the performance evaluation process. The advantage of EPS is the timely and automated management and performance evaluation of the ensemble algorithms. Therefore, the performance of EPS, for flood modeling in particular, can be improved. EPSs may use multiple fast-learning or statistical algorithms as classifier ensembles, e.g., ANNs, MLP, DTs, rotation forest (RF) bootstrap, and boosting, allowing higher accuracy and robustness. The subsequent ensemble prediction systems can be used to quantify the probability of floods, based on the prediction rate used in the event [142,143,144]. Therefore, the quality of ML ensembles can be calculated based on the verification of probability distribution. Ouyang et al [145] and Zhang et al. [146] presented a review of the applications of ensemble ML methods used for floods. EPSs were demonstrated to have the capability for improving model accuracy in flood modeling [140-146]

To improve the accuracy of import data and to achieve better dataset management, the ensemble mean was proposed as a powerful approach coupled with ML methods [140,141]. Empirical mode decomposition (EMD) [142], and ensemble EMD (EEMD) [143] are widely used for flood prediction [144]. Nevertheless, EMD-based forecast models are also subject to a number of drawbacks [145]. The literature includes numerous studies on improving the performance of decomposition and prediction models in terms of additivity and generalization ability [146].

### *3.8. Classification of ML Methods and Applications*

The most popular ML modeling methods for flood prediction were identified in the previous section, including ANFIS, MLP, WNN, EPS, DT, RF, CART, and ANN. Figure 3 presents the major ML methods used for flood prediction, and the number of corresponding articles in the literature over the last decade. This figure was designed to communicate to the readers which ML methods increased in popularity among hydrologists for flood modeling within the past decade.

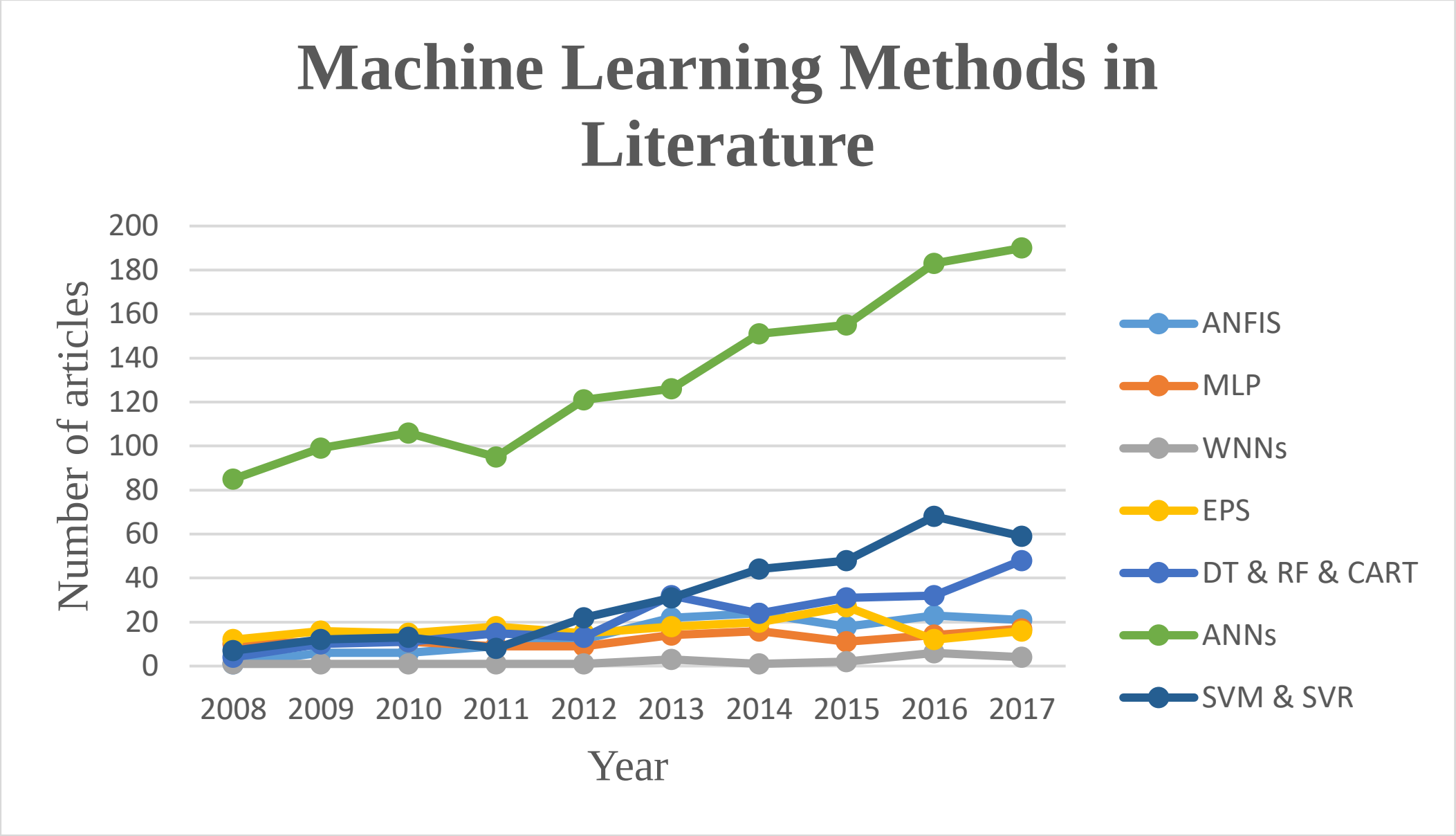

**Figure 3.** Major ML methods used for flood prediction in the literature. Reference year: 2008 (source: Scopus).

Considering the ML methods for application to floods, it is apparent that ANNs, SVMs, MLPs, DTs, ANFIS, WNNs, and EPSs are the most popular. These ML methods can be categorized as single and hybrid methods. In addition to the fundamental hybrid ML methods, i.e., ANFIS, WNNs, and basic EPSs, several different research strategies for obtaining better prediction evolved [137]. The strategies involved developing hybrid ML models using soft computing techniques, statistical methods, and physical models rather than individual ML approaches, whereby the extra components complement each other with respect to their drawbacks and shortcomings. The success of such hybrid approaches motivated the research community to explore more advanced hybrid models. Figure 4 presents the progress of single vs. hybrid ML methods for flood prediction in the literature over the past decade. The figure shows an apparent continuous increase and notable progress in using novel hybrid methods. Through Figure 4, the taxonomy of our research was justified, based on distinguishing hybrid and single ML prediction models.

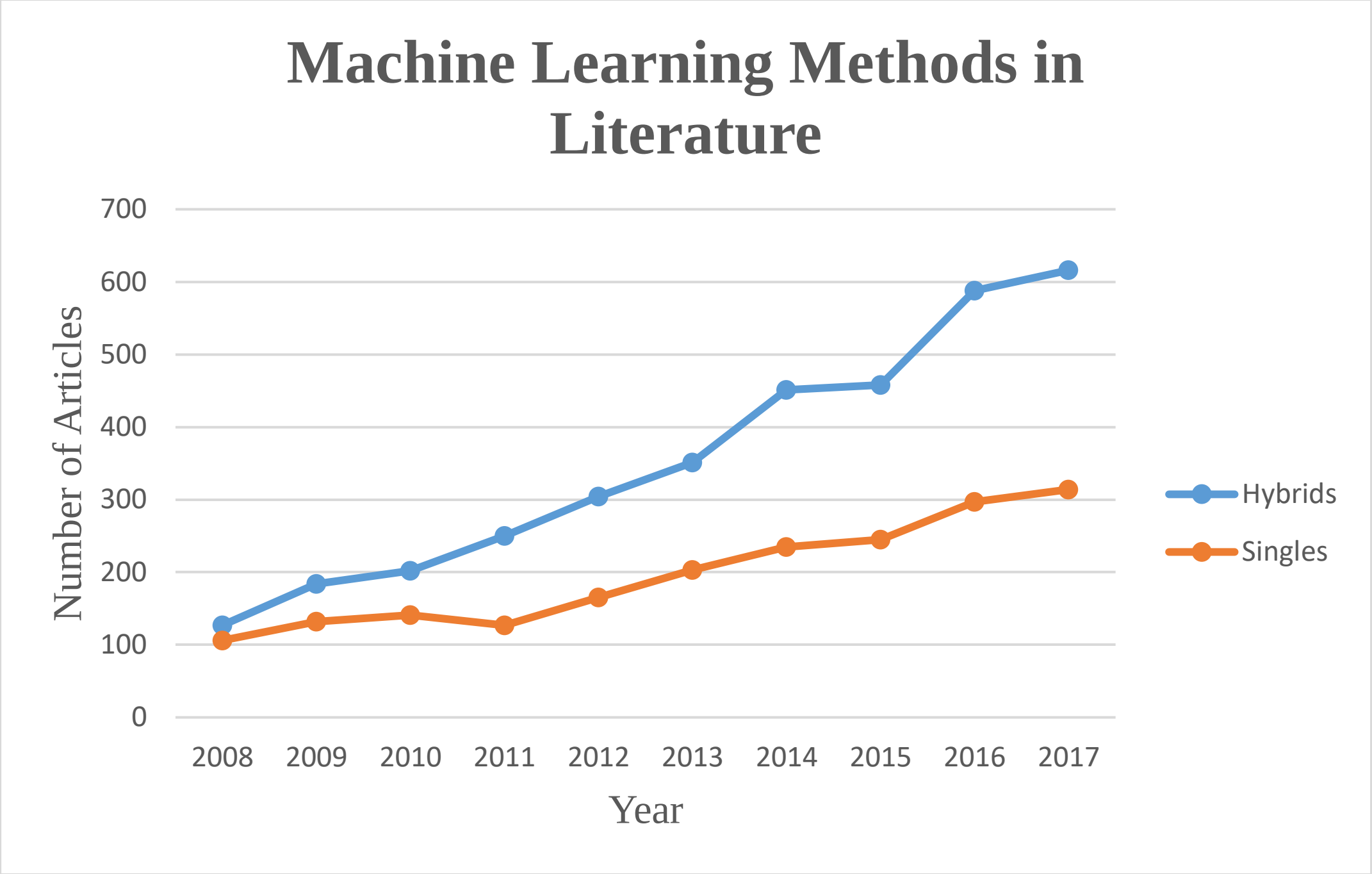

**Figure 4.** The progress of single vs. hybrid ML methods for flood prediction in the literature. Reference year: 2008 (source: Scopus).

Furthermore, the types of prediction are often studied with different lead-time predictions due to the flood. Real-time, hourly, daily, weekly, monthly, seasonal, annual, short-term, and long-term are the terms most often used in the literature. Real-time prediction is concerned with anywhere between few minutes and an hour preceding the flood. Hourly predictions can be 1–3 h ahead of the flood forecasting lead time or, in some cases, 18 h or 24 h. Daily predictions can be 1–6 days ahead of the forecast. Monthly forecasts can be, for instance, up to three months. In hydrology, the definitions of short-term and long-term in studying the different phenomena vary. Short-term predictions for floods often refer to hourly, daily, and weekly predictions, and they are used as warning systems. On the other hand, long-term predictions are mostly used for policy analysis purposes. Furthermore, if the prediction leading time to flood is three days longer than the confluence time, the prediction is considered to be long-term [37,58]. From this perspective, in this study, we considered a lead time greater than a week as a long-term prediction. It was observed that the characteristics of the ML methods used varied significantly according to the period of prediction. Thus, dividing the survey on the basis of short-term and long-term was essential.

Here, it is also worth emphasizing that, in this paper, the prediction lead-time was classified as "short-term" or "long-term". Although flash floods happen in a short period of time with great destructive power, they can be predicted with either "short-term" or "long-term" lead times to the actual flood. In fact, this paper is concerned with the lead times instead of the duration or type of flood. If the lead-time prediction to a flash flood was short-term, then it was studied as a short-term lead time. However, sometimes flash floods can be predicted with long lead times. In other words, flash floods might be predicted one month ahead. In this case, the prediction was considered as long-term. Regardless of the type of flood, we only focused on the lead time.

In this study, the ML methods were reviewed using two classes—single methods and hybrid methods. Figures 5 and 6 represent the taxonomy of the research.

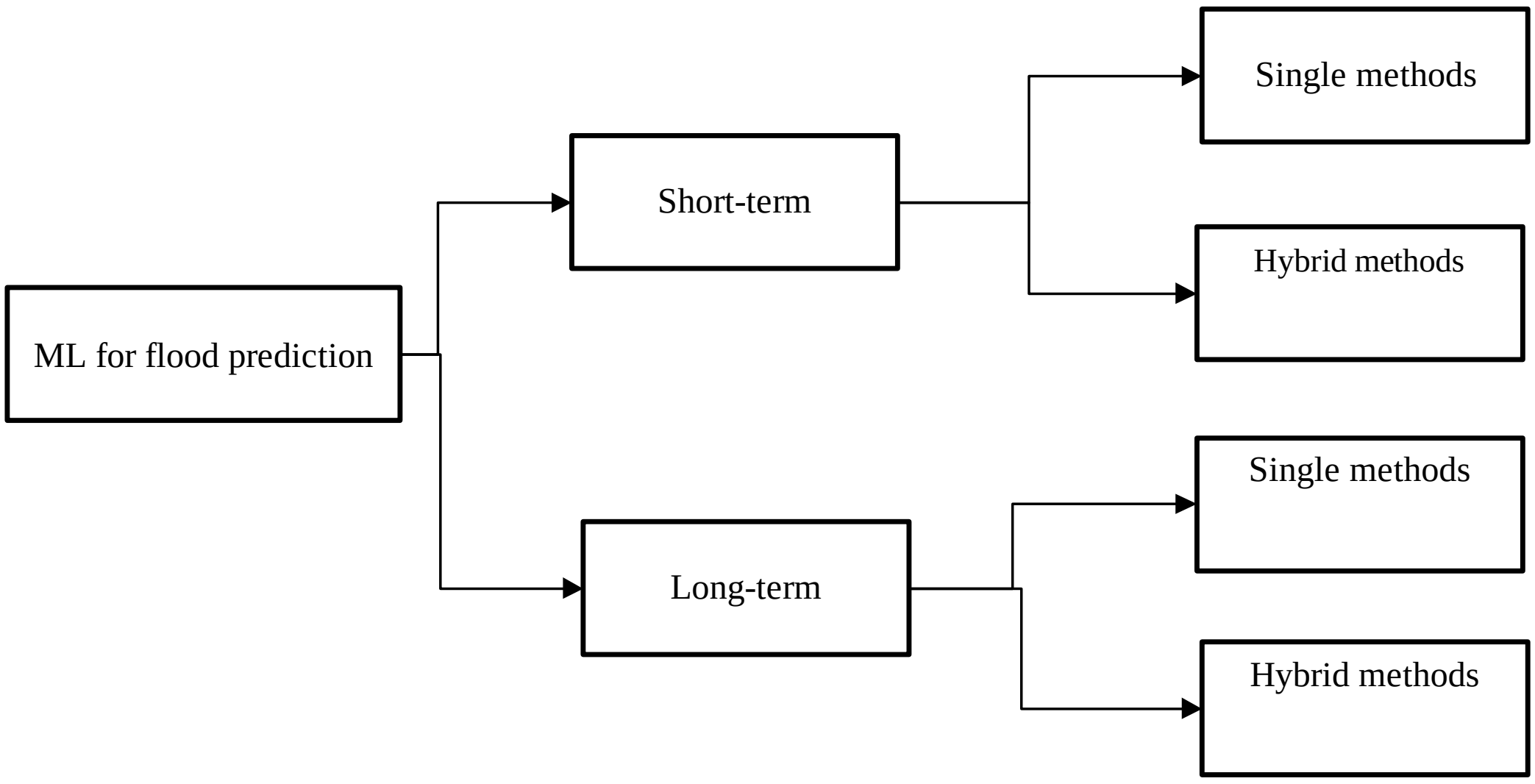

**Figure 5.** Taxonomy of the survey—ML methods for flood prediction.

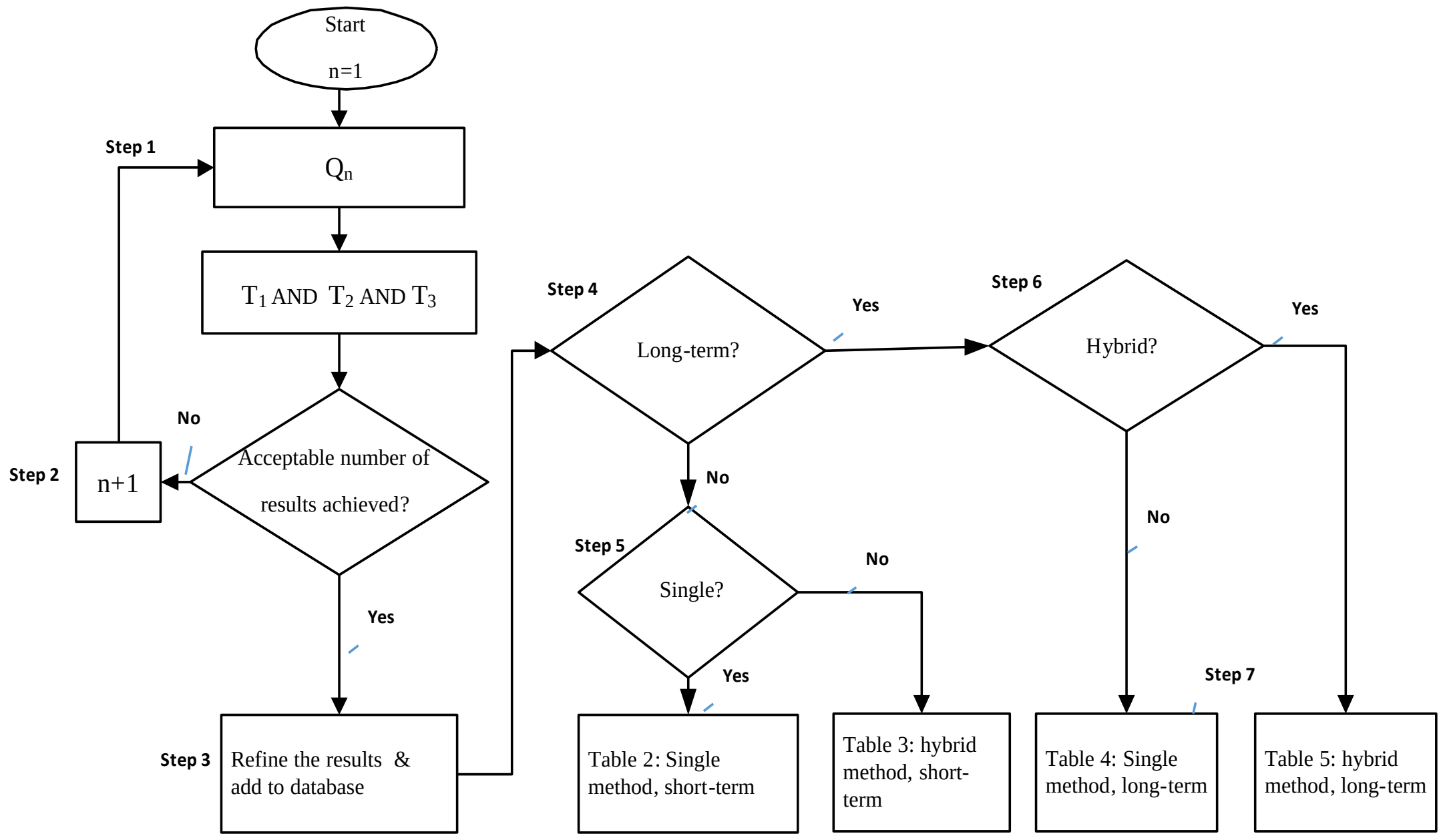

**Figure 6.** Taxonomy of the survey.

Step 1 involved running the queries one by one; step 2 involved checking the results of the search, and initiating the next search; step 3 involved identifying the comparative studies on ML models of prediction, refining the results and building the database; step 4 involved identifying whether it was a long-term or short-term prediction; steps 5 and 6 involved identifying if it was a single or hybrid method, constructing Table 1, and step 7 involved constructing the other Tables. The four tables provide the list of studies on different prediction techniques, which entail the organized comprehensive surveys of the literature.

## 4. Short-Term Flood Prediction with ML

Short-term lead-time flood predictions are considered important research challenges, particularly in highly urbanized areas, for timely warnings to residences so to reduce damage [146]. In addition, short-term predictions contribute highly to water recourse management. Even with the recent improvements in numerical weather prediction (NWP) models, artificial intelligence (AI) methods, and ML, short-term prediction remains a challenging task [147-152]. This section is divided into two subsections—single and hybrid methods of ML—to individually investigate each group of methods.

### *4.1. Short-Term Flood Prediction Using Single ML Methods*

To gain insight into the performance of ML methods, a comprehensive comparison was required to investigate ML methods. Table 1 presents a summary of the major ML methods, i.e., ANNs, MLP, nonlinear autoregressive network with exogenous inputs (NARX), M5 model trees, DTs, CART, SVR, and RF, followed by a comprehensive performance comparison of single ML methods in short-term flood prediction. A revision and discussion of these methods follow so as to identify the most suitable methods presented in the literature.

**Table 1.** Short-term predictions using single machine learning (ML) methods.

| Modeling Technique | Reference | Flood Resource Variable | Prediction Type | Region |
|---|---|---|---|---|
| ANN vs. statistical | [1] | Streamflow and flash food | Hourly | USA |
| ANN vs. traditional | [44] | Water and surge level | Hourly | Japan |
| ANN vs. statistical | [149] | Flood | Real-time | UK |
| ANN vs. statistical | [150] | Extreme flow | Hourly | Greece |
| FFANN vs. ANN | [151] | Water level | Hourly | India |
| ANN vs. T–S | [4] | Flood | Hourly | India |
| ANN vs. AR | [153] | Stage level and streamflow | Hourly | Brazil |
| MLP vs. Kohonen NN | [154] | Flood frequency analysis | Long-term | China |
| BPANN | [155] | Peak flow of flood | Daily | Canada |
| BPANN vs. DBPANN | [156] | Rainfall–runoff | Monthly and daily | China |
| BPANN | [157] | Flash flood | Real-time | Hawaii |
| BPANN | [158] | Runoff | Daily | India |
| ELM vs. SVM | [159] | Streamflow | Daily | China |
| BPANN vs. NARX | [160,161] | Urban flood | Real-time | Taiwan |
| FFANN vs. Functional ANN | [162] | River flows | Real-time | Ireland |
| Recurrent NN vs. Z–R relation | [163] | Rainfall prediction | Real-time | Taiwan |
| ANN vs. M5 model tree | [164] | Peak flow | Hourly | India |
| NBT vs. DT vs. Multinomial regression | [165] | Flash flood | Real-time, hourly | Austria |
| DTs vs. NBT vs. ADT vs. LMT, and REPT | [166] | Flood | Hourly/daily | Iran |
| MLP vs. MLR | [167,168] | River flow and rainfall–runoff | Daily | Algeria |
| MLP vs. MLR | [98] | River runoff | Hourly | Morocco |
| MLP vs. WT vs. MLR vs. ANN | [169] | River flood forecasting | Daily | Canada |
| ANN vs. MLP | [170] | River level | Hourly | Ireland |
| MLP vs. DT vs. CART vs. CHAID | [171] | Flood during typhoon | Rainfall–runoff | China |
| SVM vs. ANN | [120] | Rainfall extreme events | Daily | India |
| ANN vs. SVR | [48] | Flood | Daily | Canada |
| RF vs. SVM | [69] | Rainfall | Hourly | Taiwan |

Kim and Barros [148] modified an ANN model to improve flood forecasting short-term lead time through consideration of atmospheric conditions. They used satellite data from the ISCCP-B3 dataset [172]. This dataset includes hourly rainfall from 160 rain gauges within the region. The ANN was reported to be considerably more accurate than the statistical models. In another similar work, Reference [44] developed an ANN forecast model for hourly lead time. In their study, various datasets were used, consisting of meteorological and hydrodynamic parameters of three typhoons. Testing of the ANN forecast models showed promising results for 5-h lead time. In another attempt, Danso-Amoako [1] provided a rapid system for predicting floods with an ANN. They provided a reliable forecasting tool for rapidly assessing floods. An $R^2$ value of 0.70 for the ANN model proved that the tool was suitable for predicting flood variables with a high generalization ability. The results of [149] provides similar conclusions. Furthermore, Panda, Pramanik, and Bala [151] compared the accuracy of ANN with FFANN, and the results were benchmarked with the physical model of MIKE 11 for short-term water level prediction. This dataset includes the hourly discharge and water level between 2006 and 2009. The data of the year 2006 was used for testing root-mean-square error (RMSE). The results indicated that the FFANN performed faster and relatively more accurately than the ANN model. Here, it is worth mentioning that the overall results indicated that the neural networks were superior compared to the one-dimensional model MIKE 11. Nevertheless, there were great advancements reported in the implementation of two-dimensional MIKE 11 [8].

Kourgialas, Dokou, and Karatzas [150] created a modeling system for the prediction of extreme flow based on ANNs 3 h, 12 h, and 19 h ahead of the flood. They analyzed five years of hourly data to investigate the ANN effectiveness in modeling extreme flood events. The results indicated it to be highly effective compared to conventional hydrological models. Lohani, Goel, and Bhatia [4] improved the real-time forecasting of rainfall–runoff of foods, and the results were compared to the T–S fuzzy model and the subtractive-clustering-based T–S (TSC-T–S) fuzzy model. They, however, concluded that the fuzzy model provided more accurate predictions with longer lead time. The hourly rainfall data from 1989 to 1995 of a gauge site, in addition to the rainfall during a monsoon, was used. Pereira Filho and dos Santos [153] compared the AR model with an ANN in simulating forecast stage level and streamflow. The dataset was created from independent flood events, radar-derived rainfall, and streamflow rain gauges available between 1991 and 1995. The AR and ANN were employed to model short-term flood in an urban area utilizing streamflow and weather data. They showed that the ANN performed better in its verification and it was proposed as a better alternative to the AR model.

Ahmad and Simonovic [155] used a BPNN for predicting peak flow utilizing causal meteorological parameters. This dataset included daily discharge data for 1958–1997 from gauging stations. BPNN proved to be a fast and accurate approach with the ability of generalization for application to other locations with similar rivers. Furthermore, to improve the simulation of daily streamflow using BPNN, Reference [156] used division-based backpropagation to obtain satisfying results. The raw data of local evaporation and rainfall gauges of six years were used for the short-term flood prediction of a streamflow time series. The dataset of one decade from 1988 was used for training and the dataset of five subsequent years was used for testing. The BPNN model provided promising results; however, it lacked efficiency in using raw data for the time-series prediction of streamflow. In addition, Reference [157] showed the application of BPNN for assessing flash floods using measured data. This dataset included 5-min-frequency water quality data and 15-min-frequency rainfall data of 20 years from two rain gauge stations. Their experiments introduced ANN models as simple ML methods to apply, while simultaneously requiring expert knowledge by the user. In addition, their ANN prediction model showed great ability to deal with a noisy dataset. Ghose [158] predicted the daily runoff using a BPNN prediction model. The data of daily water level of two years from 2013–2015 were used. The accurate BPNN model was reported with an efficiency of 96.4% and an $R^2$ of 0.94 for flood prediction.

Pan, Cheng, and Cai [159] compared the performances of ELM and SVM for short-term streamflow prediction. Both methods demonstrated a similar level of accuracy. However, ELM was suggested as a faster method for parameter selection and learning loops. Reference [154] also conducted a comparison between fuzzy c-means, ANN, and MLP using a common dataset of sites to investigate ML method efficiency and accuracy. The MLP and ANN methods were proposed as the best methods. Chang, Chen, Lu, Huang, and Chang [160] and Reference [161] modeled multi-step urban flood forecasts using BPNN and a nonlinear autoregressive network with exogenous input (NARX) for hourly forecasts. The results demonstrated that NARX worked better in short-term lead-time prediction compared to BPNN. The NARX network produced an average $R^2$ value of 0.7. This study suggested that the NARX model was effective in urban flood prediction. Furthermore, Valipour et al. [24] showed how the accuracy of ANN models could be increased through integration with autoregressive (AR) models.

Bruen and Yang [162] modeled real-time rainfall–runoff forecasting for different lead times using FFNN, ARMA, and functional networks. Here, functional networks [173] were compared with an FFNN model. The models were tested using a storm time-series dataset. The result was that functional networks allowed quicker training in the prediction of rainfall–runoff processes with different lead times. The models were able to predict floods with short lead times. Reference [164] estimated water level–discharge using M5 trees and ANN. This dataset was collected from the period of 1990 to 1998, and the inputs were supplied by computing the average mutual information. The ANN and M5 model tree performed similar in terms of accuracy. Reference [166] tested four DT models, i.e., alternating decision trees (ADTs), reduced-error pruning trees (REPTs), logistic model trees (LMTs), and NBTs, using a dataset of 200 floods. The ADT model was reported to perform better for flash-flood prediction for a speedy determination of flood-susceptible areas. In other research, Reference [165] compared the performance of an NBT and DT prediction model, using geomorphological disposition parameters. Both models and their hybrids were compared in terms of prediction accuracy in a catchment. The advanced DTs were found to be promising for flood assessment in prone areas. They concluded that an independent dataset and benchmarking of other ML methods were required for judgment of the accuracy and efficiency of the method. Reference [171] worked on a dataset including more than 100 tropical cyclones (TCs) affecting a watershed for the hourly prediction of precipitation. The performances of MLP, CART, CHAID, exhaustive CHAID, MLR, and CLIM were compared. The evaluation results showed that MLP and DTs provided better prediction. Reference [163] applied a dynamic ANN, as well as a Z–R relation approach for constructing a one-hour-ahead prediction model. This dataset included three-dimensional radar data of typhoon events and rain gauges from 1990 to 2004, including various typhoons. The results indicated that the ANN performed better.

Aichouri, Hani, Bougherira, Djabri, Chaffai, and Lallahem [167] implemented an MLP model for flood prediction, and compared the results with the traditional MLR model. The rainfall–runoff daily data from 1986 to 2003 were used for model building. The results and comparative study indicated that the MLP approach performed with better yield for river rainfall–runoff. In a similar research, Reference [98] modeled and predicted the river rainfall–runoff relationship through training six years of collected daily rainfall data using MLP and MLR (1990 to 1995). Furthermore, the data of 1996 were used for testing to select the best performing network model. The $R^2$ values for the ANN and MLR models were 0.888 and 0.917, respectively, showing that the MLP approach gave a much better prediction than MLR. Reference [169] proposed a number of data-based flood predictions for daily stream flows models using MLP, WT, MLR, ARIMA, and ANN. This dataset included two time series of streamflow and a meteorological dataset including records from 1970 to 2001. The results showed that MLP, WT, and ANN performed generally better. However, the proposed WT prediction model was evaluated to be not as accurate as ANN and MLP for a one-week lead time. Reference [170] designed optimal models of ANN and MLP for the prediction of river level. This study indicated that an optimization tool for the ANN network can highly improve prediction quality. The candidate inputs included river levels and mean sea-level pressure (SLP) for the period of 2001–2002. The MLP was identified as the most accurate model for short-term river flood prediction.

Nayak and Ghosh [120] used SVM and ANN to predict hourly rainfall–runoff using weather patterns. A model of SVM classifier for rainfall prediction was used and the results were compared to ANN and another advanced statistical technique. The SVM model appeared to predict extreme floods better than the ANN. Furthermore, the SVM model proved to function better in terms of uncertainty. Gizaw and Gan [48] developed SVR and ANN models for creating RFFA to estimate regional flood quantiles and to assess climate change impact. This dataset included daily precipitation data obtained from gauges from 1950 to 2016. RMSE and $R^2$ were used for the evaluation of the models. The SVR model estimated regional flood more accurately than the ANN model. SVR was reported to be a suitable choice for predicting future flood under the uncertainty of climate change scenarios [118]. In a similar attempt, Reference [69] provided effective real-time flood prediction using a rainfall dataset measured by radar. Two models of RF and SVM were developed and their prediction performances were compared. Their performance comparison revealed the effectiveness of SVM in real-time flood forecasting.

Table 2 represents a comparative analysis of single ML models for the prediction of short-term floods, considering the complexity of the algorithm, ease of use, running speed, accuracy, and input dataset. This table was created based on the revisions that were made on the articles of Table 1 and also the accuracy analysis of Figure 3, where the values of $R^2$ and RMSE of the single ML methods were considered. The quality of ML model prediction, in terms of speed, complexity, accuracy, and ease of use, was continuously improved through using ensembles of ML methods, hybridization of ML methods, optimization algorithms, and/or soft computing techniques. This trend of improvement is discussed in detail in the discussion.

**Table 2.** Comparative analysis of single ML models for the prediction of short-term floods.

| Modeling Technique | Complexity of Algorithm | Ease of Use | Speed | Accuracy | Input Dataset |
|---|---|---|---|---|---|
| ANN | High | Low | Fair | Fair | Historical |
| BPANN | Fairly high | Low | Fairly high | Fairly high | Historical |
| MLP | Fairly high | Fair | High | Fairly high | Historical |
| ELM | Fair | Fairly high | Fairly high | Fair | Historical |
| CART | Fair | Fair | Fair | Fairly high | Historical |
| SVM | Fairly high | Low | Low | Fair | Historical |
| ANFIS | Fair | Fairly high | Fair | Fairly high | Historical |

### *4.2. Short-Term Flood Prediction Using Hybrid ML Methods*

To improve the quality of prediction, in terms of accuracy, generalization, uncertainty, longer lead time, speed, and computation costs, there is an ever increasing trend in building hybrid ML methods. These hybrid methods are numerous, including more popular ones, such as ANFIS and WNN, and further novel algorithms, e.g., SVM–FR, HEC–HMS–ANN, SAS–MP, SOM–R-NARX, wavelet-based NARX, WBANN, WNN–BB, RNN–SVR, RSVRCPSO, MLR–ANN, FFRM–ANN, and EPSs. Table 3 presents these methods; a revision of the methods and applications follows along with a discussion on the ML methods.

**Table 3.** Short-term flood prediction using hybrid ML methods.

| Modeling Technique | Reference | Flood resource Variable | Prediction Type | Region |
|---|---|---|---|---|
| ANFIS vs. ANN | [174] | Flash floods | Real-time | Spain |
| ANFIS vs. ANN | [175,176] | Water level | Hourly | Taiwan |
| ANFIS vs. ANN | [46] | Watershed rainfall | Hourly | Taiwan |
| ANFIS vs. ANN | [67] | Flood quantiles | Real-time | Canada |
| ANN vs. ANFIS | [177] | Daily flow | Daily | Iran |
| CART vs. ANFIS vs. MLP vs. SVM | [134] | Sediment transport | Daily | Iran |
| MLP vs. GRNNM vs. NNM | [96] | Flood prediction | Daily | Korea |
| SVM-FR vs. DT | [178] | Rainfall–runoff | Real-time | Malaysia |
| HEC–HMS–ANN vs. HEC–HMS-SVR | [179] | Rainfall–runoff | Hourly | Taiwan |
| SAS–MP vs. W-SAS–MP | [180] | Flash flood and streamflow | Daily | Turkey |
| SOM–R-NARX vs. R-NARX | [181] | Regional flood | Hourly | Taiwan |
| Wavelet-based NARX vs. ANN, vs. WANN | [182] | Streamflow forecasting | Daily | India |
| WBANN vs. WANN vs. ANN vs. BANN | [105] | Flood | Hourly | India |
| ANN–hydrodynamic model | [183] | Flood prediction: tidal surge | Hourly | UK |
| RNN–SVR, RSVRCPSO | [184] | Flash flood: rainfall forecasting | Hourly | Taiwan |
| AME and SSNN vs. ANN | [185] | Rainfall forecasting | Hourly | Taiwan |
| Hybrid of FFNN with linear model | [186] | Flood forecasting: daily flows | Daily | India |
| FFNN vs. FBNN vs. FFRM–ANN | [187] | Flash floods | Hourly | Taiwan |
| ANN–NLPM vs. ANN | [188] | Rainfall–runoff | Daily | China |
| EPS of MLP vs. SVM vs. RF | [189] | Runoff simulations | Real-time | Germany |
| EPS of ANNs | [190] | Flood | Daily | Canada |

Jimeno-Sáez, et al. [174] modeled flash floods using ANN and ANFIS, applying a dataset collected from 14 different streamflow gauge stations. RMSE and $R^2$ were used as evaluation criteria. The results showed that ANFIS demonstrated a considerably superior ability to estimate real-time flash floods compared to ANN. Chang and Chang [175] constructed an accurate water level forecasting system based on ANFIS for 1–3 h ahead of the flood. The ANFIS successfully provided accurate water level prediction. The hourly water level of five gauges from 1971 to 2001 was used. They concluded that the ANFIS model could efficiently deal with a big dataset [176] through fast learning and reliable prediction. A further comparison showed that the ANFIS hybrid model tuned by SVR provided superior prediction accuracy and good cost-effective computation for nonlinear and real-time flood prediction. In addition, the model with human interaction could provide better performance. In another similar research, Reference [46] developed an ANFIS model based on a precipitation dataset, which provided reliable hourly predictions with an $R^2$ more than 0.85. The results were reported as highly satisfactory for the typhoon season. Reference [67] used ANFIS for ungauged sites of 151 catchments; the results were evaluated and compared to the ANN, NLR, NLR-R modes using a Jackknife procedure. The evaluation showed that the ANFIS model provided higher generalization capability compared to the NLR and ANN models. The ANFIS model implemented an efficient mechanism for forecasting the flood region, and providing insight from the data, leading to prediction. Rezaeianzadeh (2014) [177] presented a number of forecasting systems for daily flow prediction using ANN, ANFIS, MLR, and MNLR. Furthermore, the performances of the models were calculated with RMSE and $R^2$. This dataset included precipitation data from various meteorological stations. Furthermore, the evaluation showed that MNLR models with lower RMSE values had a better performance than the ANFIS, MLR, and ANN models. Furthermore, MNLR was suggested as a low-cost and efficient model for the daily prediction of flow. In a similar attempt, Choubin, Darabi et al. (2018) [133] evaluated the accuracy of ANFIS, considering three common ML modeling tools—CART, SVM, and MLP. The evaluation suggested that the CART model performed best. Therefore, CART was strongly suggested as a reliable prediction tool for hydro-meteorological datasets. Kim and Singh [96] developed three models, namely generalized regression ANN (GRNNM), Kohonen self-organizing feature maps ANN (KSOFM–NNM), and MLP, for flood prediction. Furthermore, the prediction performance was evaluated, showing that KSOFM–NNM performed accurately compared to MLP and GRNNM in forecasting flood discharge. The hybrid models, overall, were shown to overcome the difficulties when using single ANN models. Reference [178] proposed an advanced ensemble model through combining FR and SVM to build spatial modeling in flood prediction. The results were compared with DT. This dataset included an inventory map of flood prediction in various locations. To build the model, up to 100 flood locations were used for training and validation. The evaluation results showed a high success rate for the ensemble model. The results proved the efficiency, accuracy, and speed of the model in the susceptibility assessment of floods.

Young, Liu, and Wu [179] developed a hybrid physical model through integrating the HEC–HMS model with SVM and ANN for accurate rainfall–runoff modeling during a typhoon. The hybrid models of HEC–HMS–SVR and HEC–HMS–ANN had acceptable capability for hourly prediction. However, the SVR model had much better generalization and accuracy ability in runoff discharge predictions. It was concluded that the predictions of HEC–HMS were improved through ML hybridization. Reference [180] proposed SAS–MP, which is a hybrid of wavelet and season multilayer perceptron for daily rainfall prediction. The season algorithm is a novel decomposition technique used to improve data quality. The resulting hybrid model was referred to as the W-SAS–MP model. This dataset included the daily rainfall data of three decades since 1974. The W-SAS–MP model was reported as highly efficient for enhancing daily rainfall prediction accuracy and lead time.

Chang, Shen, and Chang [181] developed a hybrid ANN model for real-time forecasting of regional floods in an urban area. The advanced hybrid model of SOM–R-NARX was an integration of the NARX network with SOM. Their big dataset included 55 rainfall events of daily rainfall. The evaluation suggested that SOM–R-NARX was accurate with small values of RMSE and high $R^2$. Furthermore, compared to the cluster-based hybrid inundation model (CHIM), it provided hourly prediction accuracy. Reference [182] proposed a model of wavelet-based NARX (WNARX) for the daily forecasting of rainfalls on a dataset of gauge-based rainfall data for the period from 2000 to 2010. The prediction performance was further benchmarked with ANN, WANN, ARMAX, and NARX models, whereby WNARX was reported as superior.

Partal [110] developed a model for the daily prediction of precipitation with ANN and WNN models. In their case, WNN showed significantly better results with an average value of 0.79 at various stations. In Reference [60], they compared WNN with ANFIS for daily rainfall. The results showed that the hybrid algorithm of WNN performed better with an $R^2$ equal to 0.9 for daily lead time. Reference [105] proposed a hybrid model of wavelet, bootstrap technique, and ANN, which they called WBANN. It improved the accuracy and reliability of the ANN model short-term flood prediction. The performance of WBANN was compared with bootstrap-based ANNs (BANNs) and WNN. The wavelet decomposition significantly improved the ANN models. In addition, the bootstrap resampling produced consistent results. French, Mawdsley, Fujiyama, and Achuthan [183] proposed a novel hybrid model of ANN and a hydrodynamic model for the accurate short-term prediction of extreme storm surge water. The ANN–hydrodynamic model generated realistic flood extents and a great improvement in model accuracy. Reference [184] proposed a hybrid forecasting technique called RSVRCPSO to accurately estimate the rainfall. RSVRCPSO is an integration of RNN, SVR, and a chaotic particle swarm optimization algorithm (CPSO). This dataset was obtained from three rain gauges from the period of 1985 to August 1997, which included the data of nine typhoon events. The results suggested that the proposed model yielded better performance for rainfall prediction. The RSVRCPSO model, in comparison with SVRCPSO, resulted in less RMSE learning and testing, which gave way to superiority in prediction.

Pan et al. [185] proposed a monsoon rainfall enhancement (AME) based on ANNs, which was a hybrid form of linear regression and a state-space neural network (SSNN). The performance of the proposed model was benchmarked against the hybrid method of MLR–ANN. This dataset included the total rain, wind, and humidity measures from 1989–2008 based on 371 rain gauge stations of six typhoons. The results indicated that the method was highly robust with a better prediction accuracy in terms of $R^2$, peak discharge, and total volume. Rajurkar et al. [186] modeled rainfall–runoff by integrating ANN and a simplified linear model. Furthermore, this dataset included the daily measurements of rainfall in the period of 1963–1990. The hybrid model was found to be better for providing a theoretical forecasting representation of floods with $R^2$ equal to 0.728.

Hsu et al. [187] proposed a hybrid model from the integration of a flash-flood routing model (FFRM) and ANN, called the FFRM–ANN model, to predict hourly river stages. The ANN algorithms used in this study were the FFNN and FBNN. Data from eight typhoon events between 2004 and 2005 of rainfall and river stage pairs were selected for model training. The results indicated that the hybrid model of FFRM–ANN provided an efficient FFRM for accurate flood forecasting. The comparison of the hybrid method against each algorithm used in the study proved the effectiveness of the proposed method. Reference [188] developed a hybrid prediction model by integrating ANN and a nonlinear perturbation model (NLPM), defined as NLPM–ANN, to improve the efficiency and accuracy of rainfall–runoff prediction. The model of NLPM–ANN was benchmarked against two models of nonlinear perturbation model (LPM), and NLPM integrated with antecedent precipitation index (API) i.e., NLPM–API, on a dataset of daily rainfall–runoff in the period of 1973–1999. They reported that the NLPM–ANN worked better than the models of LPM and NLPM–API. The results of the case studies of various watersheds proved the model accuracy.

Through an EPS model, Reference [189] aimed at limiting the range of the uncertainties in runoff simulations and flood prediction. The classifier ensembles included MLP, SVM, and RF. Note that the ensemble of MLP was a novel approach in flood prediction. The proposed EPS presented a number of integrated models and simulation runs. The model validation was successfully performed using a dataset from various rain gauges of precipitation data during the 2013–2014 storm season. Using the EPS model decreased uncertainty in forecasting, which resulted in the prediction system being evaluated as reliable and robust in estimating flood duration and destructive power. In another case, Reference [190] developed an EPS model of six ANNs for daily streamflow prediction based on daily high-flow data from the storm season of 2013–2014. The proposed model had a fast development time, which also provided probabilistic forecasts to deal with uncertainties in prediction. The ensemble prediction system was reported as highly useful and robust.

*4.3. Comparative Performance Analysis*

To evaluate a reliable prediction, the accuracy, reliability, robustness, consistency, generalization, and timeliness are suggested as the basic criteria (Singh 1989). The timeliness is one of the most important criteria, and it is only achieved through using robust yet simple models. Furthermore, the performance of the prediction models is often evaluated through root-mean-square error (RMSE), mean error (ME), mean squared error (MSE), Nash coefficients (E), and $R^2$, also known as the correlation coefficient (CC). In this survey, the values of $R^2$ and RMSE were considered for performance evaluation. CC (Eq.1) and RMSE (Eq.2) can be defined as follow:

$$CC = \frac{\sum_{i=1}^{N}(x_i-\bar{x})(y_i-\bar{y})}{\sqrt{[\sum_{i=1}^{N}(x_i-\bar{x})^2][\sum_{i=1}^{N}(y_i-\bar{y})^2]}} \qquad \text{Eq.1}$$

where $x_i$ and $y_i$ are the observed and predicted values and the $i$-th residue; $\overline{x}$ and $\overline{y}$ are their means, respectively.

$$RMSE = \sqrt{\frac{\sum_{i=1}^{n}\left(X_{obs,i} - X_{model,i}\right)^2}{n}} \qquad \text{Eq.2}$$

where $X_{obs}$ defines observed variables and $X_{model}$ prediction values for year $i$, where generally $R^2 > 0.8$ is considered as an acceptable prediction. However, a lower value for RMSE suggests a better prediction. Overall, forecasting models of floods are reported as accurate if RMSE values are close to 0, and $R^2$ values are close to 1. The specific intended purpose, computational cost, and dataset would be our major consideration criteria. Furthermore, the generalization ability, speed and cost of implementation and operation, ease of use, low-cost maintenance, robustness, and accuracy of the simulation are other important criteria for evaluation of the methods.

Here, it is worth mentioning that the value of RMSE can be different across various studies. In addition, the values of RMSE in some studies were calculated for various sites. To present a fair evaluation of RMSE, we made sure that the unit of RMSE was the same, and, for the multiple RMSEs, the average was calculated. We also double-checked for any possible error. The comparative performance analysis of single and hybrid ML methods for short-term flood prediction using $R^2$ and RMSE are presented in Figures 7 and 8 respectively.

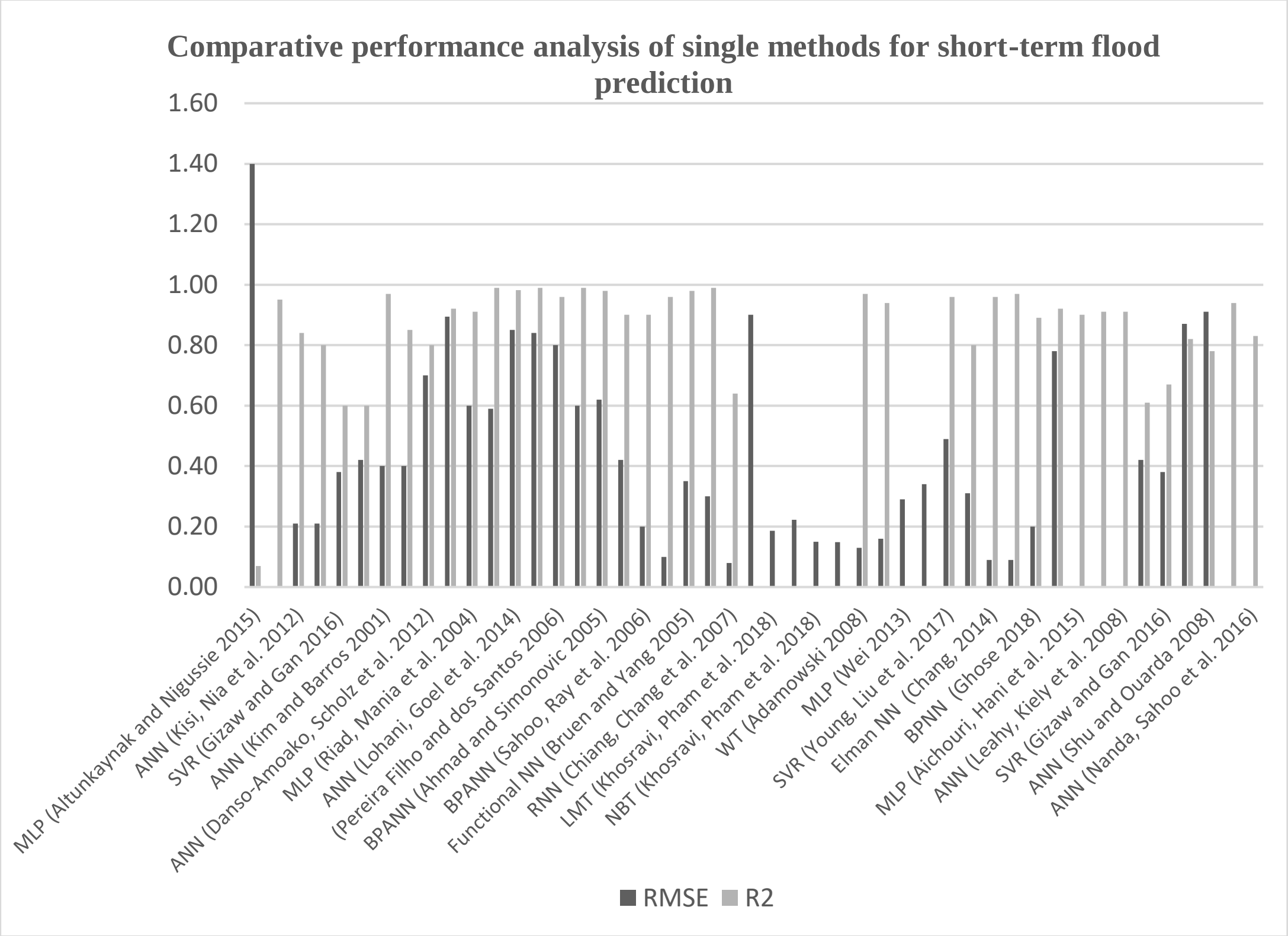

**Figure 7.** Comparative performance analysis of single methods of ML for short-term flood prediction using $R^2$ and root-mean-square error (RMSE).

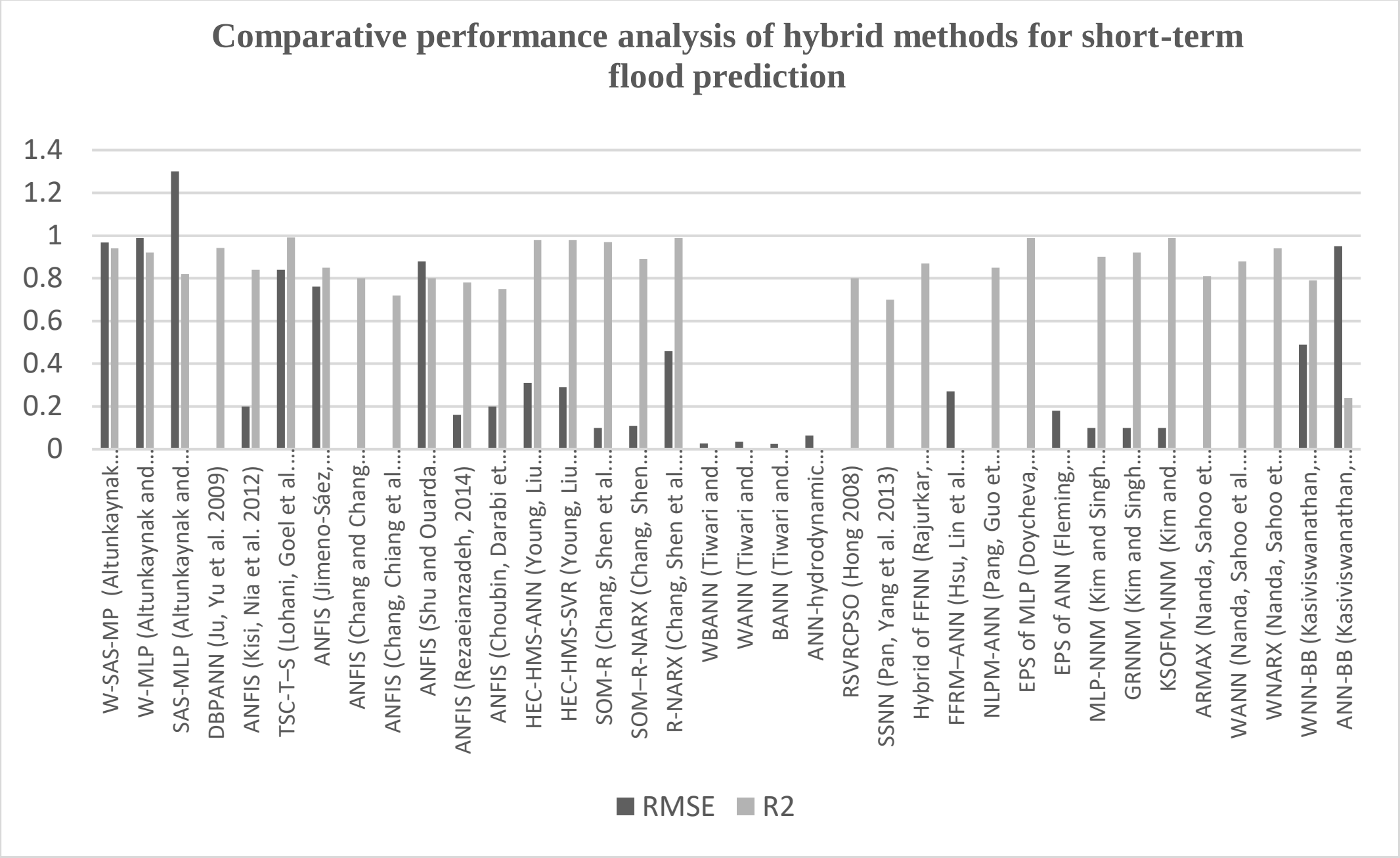

**Figure 8.** Comparative performance analysis of hybrid methods of ML for short-term flood prediction, using $R^2$ and RMSE.

Generally, ANNs are suggested as promising means for short-term prediction. Despite performing weakly in a few early studies, especially in the generalization aspect, better methodologies for higher-performance ANNs in handling big datasets yielded better results. In this context, the BPNN and functional networks are suggested as being difficult to be implemented by the user. However, the models were shown to be reasonably accurate, efficient, and fast with the ability to deal with noisy datasets. However, the NARX network performed better compared to BPNN. Nevertheless, accuracy could be enhanced through integration with autoregressive models. MLP and DTs provide equally acceptable prediction yields with ANNs. Among DTs, the ADT model provided the fastest and most accurate prediction capability in determining floods. Although not as popular as ANNs, the rotation forest (RF) and M5 model tree (MT) were reported as efficient and robust. References e.g. [69,136] proposed RF-based models that were as effective as ANNs and suitable for long lead times.

Along with ANNs, the SVM was also seen as a relatively effective ML tool for rainfall–runoff modeling and classification with better generalization ability and performance. In many cases, SVM performed even better, especially for very short lead times [122,125]. In particular, SVM-based models provided promising performances for hourly prediction. Nevertheless, the prediction ability decreased for longer lead times. This issue was addressed using the LS-SVM model, which also showed better generalization ability [127]. Generally, SVM was reported to be a suitable choice to evaluate the uncertainty in predicting hazardous flood quantiles, which revealed the effectiveness of SVM in real-time flood forecasting.

Overall, the reviewed single prediction models could provide relatively accurate short-term forecasts. However, for predictions longer than 2 h, hybrid models such as ANFIS, and WNN performed better. The performance comparisons of the ANFIS model with BPNN and AR models, with average correlation coefficients higher than 0.80, showed the superiority of ANFIS in a wide range of short-term flood prediction applications, e.g., water level, rainfall–runoff, and streamflow (for up to 24 h). ANFIS demonstrated a considerably superior ability for estimating real-time flash flood estimation compared to most ANN-based models, particularly 1–3 h ahead of flood, providing high accuracy and reliability. More advanced ANFIS hybrid models tuned by SVR provided even better prediction accuracy and good cost-effective computation for nonlinear and real-time flood prediction. Furthermore, ANFIS models presented higher generalization ability. However, by increasing the prediction lead time, $R^2$ decreased. For daily flow, MNLR was suggested with a superior performance over the ANN, ANFIS, and MLR models. In cases where hydro-meteorological data are readily available, CART was superior to ANFIS, SVM, and MLP; T–S fuzzy was also a good choice. On the other hand, WNN performed significantly better than MLP, ANNs, and ANFIS for daily predictions. For accurate longer lead-time predictions, decomposition techniques such as DWT, autoregression, and the season algorithm provided great advantages.

Overall, the novel hybrid models designed using ML, soft computing, and statistical methods, e.g., KSOFM–NNM, SOM–R-NARX, WNARX, HEC–HMS–SVR, HEC–HMS–ANN, W-SAS–MP, WBANN, RSVRCPSO, and the ANN–hydrodynamic model, were shown to overcome the drawbacks of most ML methods by enhancing the prediction accuracy and lead time, leading to more realistic flood models with even better susceptibility assessment. On the other hand, novel ensemble methods not only improved the accuracy robustness of predictions, but also contributed to limiting the range of uncertainties in models. Among the EPS methods, the ensembles of ANN, MLP, SVM, and RF showed promising results.

## 5. Long-Term Flood Prediction with ML

Long-term flood prediction is of significant importance for increasing knowledge and water resource management potential over longer periods of time, from weekly to monthly and annual predictions [191]. In the past decades, many notable ML methods, such as ANN [74], ANFIS [68,192], SVM [193], SVR [193], WNN [51], and bootstrap–ANN [51], were used for long lead-time predictions with promising results. Recently, in a number of studies (e.g., References [55,194-198]), the performances of various ML methods for long lead-time flood predictions were compared. However, it is still not clear which ML method performs best in long-term flood prediction. In this section, Tables 4 and 5 represent a summary of these investigations, and we review the performance of the ML models in dealing with long-term predictions.

### *5.1. Long-Term Flood Prediction Using Single ML Methods*

This section presents a comprehensive comparison on ML methods. Table 4 presents a summary of the major single ML methods used in long-term flood prediction, i.e., MLP, ANNs, SVM, and RT, followed by a comprehensive performance comparison. A revision and discussion of these methods follow, identifying the most suitable methods presented in the literature.

**Table 4.** Long-term flood prediction using single ML methods.

| Modeling Technique | Reference | Flood Resource Variable | Prediction Type | Region |
|---|---|---|---|---|
| ANNs | [197] | Water levels | Seasonal | Sudan |
| ANNs | [87] | Precipitation | Monthly | Australia |
| BPNNs | [199] | Heavy rainfall | Seasonal | India |
| BPNNs vs. BFGSNN | [200] | Reservoir levels | Monthly | Turkey |
| BPNN vs. MLP | [201] | Discharge | Monthly | Iran |
| ANNs vs. HBI | [202] | Stream | Weekly | Canada |
| SVM vs. ANN | [203] | Streamflow | Monthly | China |
| RT | [204] | Floodplain forests | Annually | Australia |

For seasonal flood forecasting, Elsafi [197] proposed numerous ANNs and compared the results. The water level data from different stations from 1970–1985 were selected for training, and the data from 1986–1987 were used for verification. The ANNs worked well, especially where the dataset was not complete, providing a viable choice for accurate prediction. ANNs provided the possibility of reducing the analytical costs through reducing the data analysis time that used to face in e.g., [198]. Similarly, reference [87] used ANNs to develop a prediction model for precipitation. A historical dataset of 1900–2001 of different stations was considered and the ANN model was applied to various stations to evaluate prediction performance. The authors summarized that the ANN models offered great forecasting skills for predicting long-term evapotranspiration and precipitation. Reference [202] used an ANN model for stream assessment for long-term floods. This dataset was collected from more than 100 sites of numerous flood streams. They concluded that the ANN model, compared to Hilsenhoff's biotic index (HBI), significantly improved the prediction ability using geomorphic data. However, the ANN had generalization problems. Nevertheless, the ANN in this case proved useful to water managers.

Singh [199] used a number of BPNNs to build prediction models of heavy rains and floods. This dataset included the period of 1871–2010 on a monthly time scale. The results indicated that the BPNN models were fast and robust with simple networks, which made them great for forecasting nonlinear floods. Reference [200] aimed to better analyze nonlinear floods through modeling with BPNN and local linear regression (LLR)-based models for long-term flood forecasting. This dataset included almost two decades of rainfall, outflow, inflows, evaporation, and water level since 1988. Their evaluation concluded that LLR showed better prediction than the Broyden Fletcher Goldfarb Shanno neural network (BFGSNN) model in terms of performance and accuracy with bigger values of $R^2$ and lower values of RMSE. However, BPNN outperformed the other methods with relatively good results. Among the ANN variations, [151] proposed a BPNN model as the most reliable ANN for long-term flood prediction. Reference [201] also compared the performances of ANNs with BPNN and MLP in the long-term prediction of flood discharge. Promising results were obtained when using MLP. However, generalization remained an issue.

Lin, Cheng, and Chau [203] applied an SVM model for estimating streamflow and reservoir inflow for a long lead time. To benchmark, they used ANNs and ARMA. The prediction models were built using monthly river flow discharges from the period of 1974–1998 for training, and 1999–2003 for testing. Through a comparison of model performance, SVM was demonstrated as a potential candidate for the prediction of long-term discharges, outshining the ANN. In a similar approach, Reference [205] proposed an SVM-based model for estimating soil moisture using remote-sensing data, and the results were compared to predictive models based on BPNN and MLR. Training was performed on the data of the period of 1998 to 2002, and testing used data from 2003 to 2005. The SVM model was shown to be more accurate and easier to build compared to BPNN and MLR. Reference [204], employed RT to model forest flood. Data from 2009–2012 at 50 sites were used for model building. The prediction of annual forest floods was reported through a combination of quantitative ground surveys, satellite imagery, hybrid machine learning tools, and future validation.

Table 5 presents a comparative analysis of single ML models for the prediction of long-term floods considering the complexity of algorithm, ease of use, running speed, accuracy, and input dataset. This table was created based on revisions that were made on articles of Table 4, as well as the accuracy analysis in Figure 9, where values of $R^2$ and RMSE for the single ML methods were considered. The quality of the ML model prediction, in terms of speed, complexity, accuracy, and ease of use, improved continuously through the use of ensembles of ML methods, hybridization of ML methods, optimization algorithms, and/or soft computing techniques. This trend of improvement is discussed in detail in the discussion.

**Table 5.** Comparative analysis of single ML models for the prediction of long-term floods.

| Modeling Technique | Complexity of Algorithm | Ease of Use | Speed | Accuracy | Input Dataset |
|---|---|---|---|---|---|
| ANN | Fairly high | Low | Fair | High | Historical |
| BPNN | Fairly high | Low | Fairly high | Fairly high | Historical |
| MLP | high | Fair | High | Fairly high | Historical |
| SVR | Fairly high | Low | Low | High | Historical |
| RT | Fair | Fair | Fair | Fairly high | Historical |
| SVM | Fairly high | Low | Low | High | Historical |
| M5 tree | Fair | Low | Fair | Fair | Historical |

### *5.2. Long-Term Flood Prediction Using Hybrid ML Methods*

A critical review on the long-term flood prediction using hybrid methods is presented in Table 6. Valipour, Banihabib, and Behbahani [26] used a hybrid method of autoregressive ANN integrated with sigmoid and radial activity functions. The proposed hybrid method outperformed the conventional statistical methods of ARMA and ARIMA with lower values of RMSE. They reported that ARIMA was suitable for the prediction of monthly and annual inflow, while the dynamic autoregressive ANN model with a sigmoid activity function could be used for even longer lead time. This dataset included monthly discharge from the period of 1960 to 2007.

**Table 6.** Long-term flood prediction using hybrid methods.

| **Modeling Technique** | **Reference** | **Flood Resource Variable** | **Prediction Type** | **Region** |
|---|---|---|---|---|
| Autoregressive ANN vs. ARMA vs. ARIMA | [26] | River inflow | Monthly and yearly | Iran |
| Hybrid WNN vs. M5 model tree | [206] | Streamflow water level | Monthly | Australia |
| WNN vs. ANN | [207,208] | Rainfall–runoff | Monthly | Italy |
| WNN-BB vs. WNN vs. ANN | [50] | Streamflow | Weekly and few days | Canada |
| WNN vs. ANN | [25] | Urban water | Monthly | Canada |
| WNN vs. ANN | [209] | Peak flows | Seasonal | India |
| WNN vs. ANN | [210] | Rainfall | Monthly | India |
| WARM vs. AR | [211] | Rainfall | Yearly | Thailand |
| ANFIS vs. ANNs | [212] | Rainfall | Seasonal | Australia |
| ANFIS vs. ARMA vs. ANNs vs. SVM | [213] | Discharge | Monthly | China |
| ANFIS, ANNs vs. SVM vs. LLR | [214] | Streamflow | Short-term | Turkey |
| NLPM–ANN | [215] | Flood forecasting | Yearly | China |
| M-EMDSVM vs. ANN vs. SVM | [216] | Streamflow | Monthly | China |
| SVR–DWT–EMD | [217] | Streamflow | Monthly | China |
| Surrogate modeling–ML vs. ANN–Kriging model vs. ANN–PCA | [218] | Rainfall–runoff | Yearly | USA |
| EPS of ANNs: K-NN vs. MLP vs. MLP–PLC vs. ANNE | [219] | Streamflow | Seasonal | Canada |
| EEMD–ANN vs. SVM vs. ANFIS | [220] | Runoff forecast | Monthly | China |
| WNN vs. ANN vs. WLGP | [51] | Streamflow | Monthly | Iran |

Adamowski [25] developed models based on ANN and WNN, and compared their prediction performances with statistical methods. WNN was proposed as the most accurate prediction model, as previously confirmed by Cannas et al. (2005) [207] for monthly rainfall–runoff forecasting, and also for further engineering application [208]. In a similar work, Reference [209] compared the performances of ANN and WNN for the prediction of peak flows. They also reported WNN as most reliable for simulating extreme event streams, whereby decomposition improved the results considerably. Higher levels of wavelet decomposition further improved the testing results. The statistical performance evaluation of RMSE showed considerable improvement in the testing results. Venkata Ramana [210] also combined the wavelet technique with ANN for long-term flood prediction. They considered 74 years of data for the period of 1901 to 1975. A dataset of 44 years was used for calibration, and the remainder was used for validation of the model. Their results showed a relatively lower performance for ANNs compared WNN models in modeling rainfall–runoffs. Cannas et al. [207] proposed WNN for monthly rainfall–runoff prediction, which showed significant improvement over ANNs. In a similar attempt, Kasiviswanathan, He, Sudheer, and Tay [50] used WNN and WNN–BB, which is an ensemble of WNN utilizing the block bootstrap (BB) sampling technique, to identify a robust modeling approach among ANN and WNN, by assessing accuracy and precision. This dataset included measurements from 1912 to 2013 at several flow gauge stations. The results suggested WNN–BB as a robust model for long-term streamflow prediction for longer lead times of up to one year. Tantanee et al. [211] proposed a hybrid of wavelet and autoregressive models, called WARM, which performed more effectively for long lead times. Prasad [206] proposed another similar hybrid model with the integration of WNN and iterative input selection (IIS). The hybrid model was called IIS–W-ANN, and was benchmarked with the M5 model tree. Their dataset included streamflow water level measurements from 40 years. The IIS–W-ANN hybrid model outperformed the M5 tree. This study advocated that the novel IIS–W-ANN method should be considered as an excellent flood forecasting model. Nevertheless, the model could be further optimized for better performance using optimization methods introduced in references [221–225]. In fact, such optimizers can complement IIS–W-ANN for fine-tuning the hidden-layer weights and biases for better prediction. Mekanik [212] used ANFIS to forecast seasonal rainfall. A comparison of the performance and accuracy of the ANN model and a physical model showed promising results for ANFIS. Rainfall measurements of 1900–1999 were used for training and validation, and the following decade was used for testing. The results showed that ANFIS outperformed the ANN models in all cases, comparable to Predictive Ocean Atmosphere Model for Australia (POAMA), and better than climatology. Furthermore, the study demonstrated the accuracy of ANFIS compared to global climate models. In addition, the study suggested ANFIS as an alternative tool for long-term predictions. ANFIS was reported as being easy to implement with low complexity and minimal input requirements, as well as less development time. Reference [213] compared the performances of ANFIS, ANNs, and SVM. This dataset included monthly flow data from 1953 to 2004, where the period of 2000–2004 was used for validation. ANFIS and SVM were evaluated as being better for long-term predictions. References [224,226] compared the performances of ANFIS, ANNs, and SVM for the monthly prediction of floods. The comparison results indicated that the ML models provided more accuracy than the statistical models in predicting streamflow. Furthermore, ANN and ANFIS presented more accuracy vs. SVM. However, for low-flow predictions, the SVM and ANN models outperformed ANFIS. Reference [215] proposed a modified variation of a hybrid model of NLPM–ANN to predict wetness and flood. To do so, the seasonal rainfall and wetness data of various stations were considered. The NLPM–ANN model was reported as being significantly superior to the models of previous studies. In another hybrid model, Reference [216] investigated the performance of a modified EMD–SVM (M-EMDSVM) model for long lead times, and comparedits accuracy with ANN and SVM models. The M-EMDSVM model was created through modification of EMD–SVM. The evaluation results showed that the M-EMDSVM model was a better alternative to ANN, SVM, and EMD–SVM models for long lead-time streamflow prediction. The M-EMDSVM model also presented better stability, representativeness, and precision.

Zhu, Zhou, Ye, and Meng [217] contributed to the integration of ML with time-series decomposition to predict monthly streamflow through estimation and comparison of accuracy of a number of models. For that matter, they integrated SVM with discrete wavelet transform (DWT) and EMD. The hybrid models were called DWT–SVR and EMD–SVR. The results indicated that decomposition improved the accuracy of streamflow prediction, yet DWT performed even better. Further comparisons of SVR, EMD–SVR, and DWT–SVR models showed that EMD and DWT were significantly more accurate than SVR for monthly streamflow prediction.

Araghinejad [219] presented the applicability of ensembles for probabilistic flood prediction in real-life cases. He utilized the K-nearest neighbor regression for the purpose of combining individual networks and improving the performance of prediction. As an EPS of ANNs, the hybrid model of K-NN was proposed to increase the generalization ability of neural networks, and was further compared with the results using MLP, MLP–PLC, and ANN. The hourly water level data of the reservoir from 132 typhoons in the period of 1971–2001 were used. The proposed EPS had a promising ability of generalization and prediction accuracy.

Bass and Bedient [218] proposed a hybrid model of surrogate–ML for long-term flood prediction suitable for TCs. The methods used included ANN integrated with principal component analysis (PCA), Kriging integrated with PC, and Kriging. The models were reported as efficient and fast to build. The results demonstrated that the methodology had an acceptable generalization ability suitable for urbanized and coastal watersheds. Reference [220] contributed to improving decomposition ensemble prediction models by developing an EEMD–ANN model for monthly prediction. The performance comparison with SVM, ANFIS, and ANNs showed a significant improvement in accuracy.

Ravansalar [51] compared the performances of the prediction models of WNN, ANN, and a novel hybrid model called wavelet linear genetic programming (WLGP) in dealing with the long-term prediction of streamflow. The results showed an accuracy of 0.87 for the WLGP model. The comparison of the performance evaluation showed that WLGP significantly increased the accuracy for the monthly approximation of peak streamflow.

## 6. Comparative Performance Analysis and Discussion

In this section, the comparative performance analysis of ML methods for long-term prediction is presented. Figure 9 represents the values of RMSE and $R^2$ for single methods of ML, where ANNs, SVMs, and SVRs show better results. Figure 10 represents the values of RMSE and $R^2$ for hybrid methods of ML, where decomposition and ensemble methods outperformed the more traditional methods.

ANNs are the most widely used ML method due to their accuracy, high fault tolerance, and powerful parallel processing in dealing with complex flood functions, especially where datasets are not complete. However, generalization remains an issue with ANN. In this context, ANFIS, MLP, and SVM performed better than ANNs. However, wavelet transforms were reported to be useful for decompositions of original time series, improving the ability of most ML methods by providing insight into datasets on various resolution levels as appropriate data pre-processing. For instance, WNNs generally produce more consistent results compared to traditional ANNs.

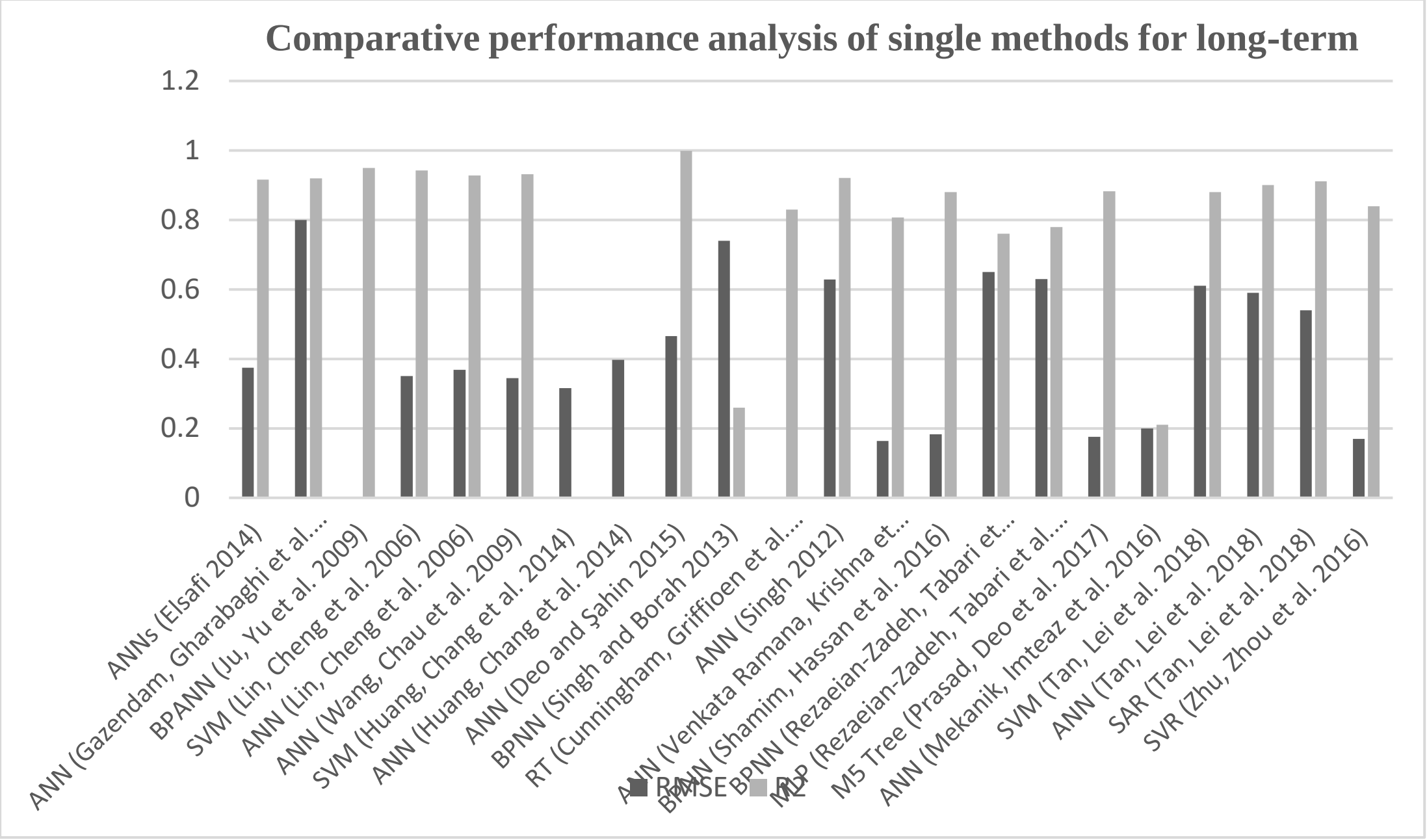

**Figure 9.** Comparative performance analysis of single methods of ML for long-term prediction.

Either in short-term [227] or long-term rainfall–runoff modeling [50], overall, the accuracy, precision, and performance of most decomposed ML algorithms (e.g., WNN) were reported as better than those which were trained using un-decomposed time series. However, despite the achievement of WNNs, the predictions were not satisfactory for long lead times. To increase the accuracy of the longer-lead-time predictions up to one year, novel hybrids such as WARM, which is a hybrid of WNN and an autoregressive model, and wavelet multi-resolution analysis (WMRA) were proposed. In other cases, it was seen that the performance of models improved greatly through decomposition to produce cleaner inputs. For example, wavelet–neuro-fuzzy models [228] were significantly more accurate and faster than single ANFIS and ANNs. However, with an increase in the lead time, the uncertainty in prediction increased. Thus, the evaluation of model precision should come into consideration in future studies.

Data decomposition methods, e.g., autoregressive, wavelet transforms, wavelet–autoregressive, DWT, IIS, and EMD, contributed highly to developing hybrid methods for longer prediction lead time, good stability, great representativeness, and higher accuracy. These data decomposition methods were integrated with ANNs, SVM, WNN, and FR, and they are expected to gain more popularity among researchers. The other trend in improvement of prediction accuracy and generalization capability involves EPS. In fact, recent ensemble methods contributed to good improvements in speed, accuracy, and generalization. The EPS of ANNs and WNNs, using BB sampling, genetic programming, simple average, stop training, Bayesian, data fusion, regression, and other soft computing techniques, showed promising results and better performances than traditional ML methods. In ensembles, however, it is noted that human decision as the input variable provided superior performance than models without this important input. However, the most significant hybrid models were novel decomposition–ensemble prediction models suitable for monthly prediction. Their performance comparisons with SVM, ANFIS, and ANNs showed significant improvements in accuracy and generalization. Figure 10 represents the comparative performance analysis of hybrid methods of ML for short-term prediction. Here, it is also worth mentioning the importance of further signal processing techniques (e.g., Reference [228]) for both long-term and short-term floods.

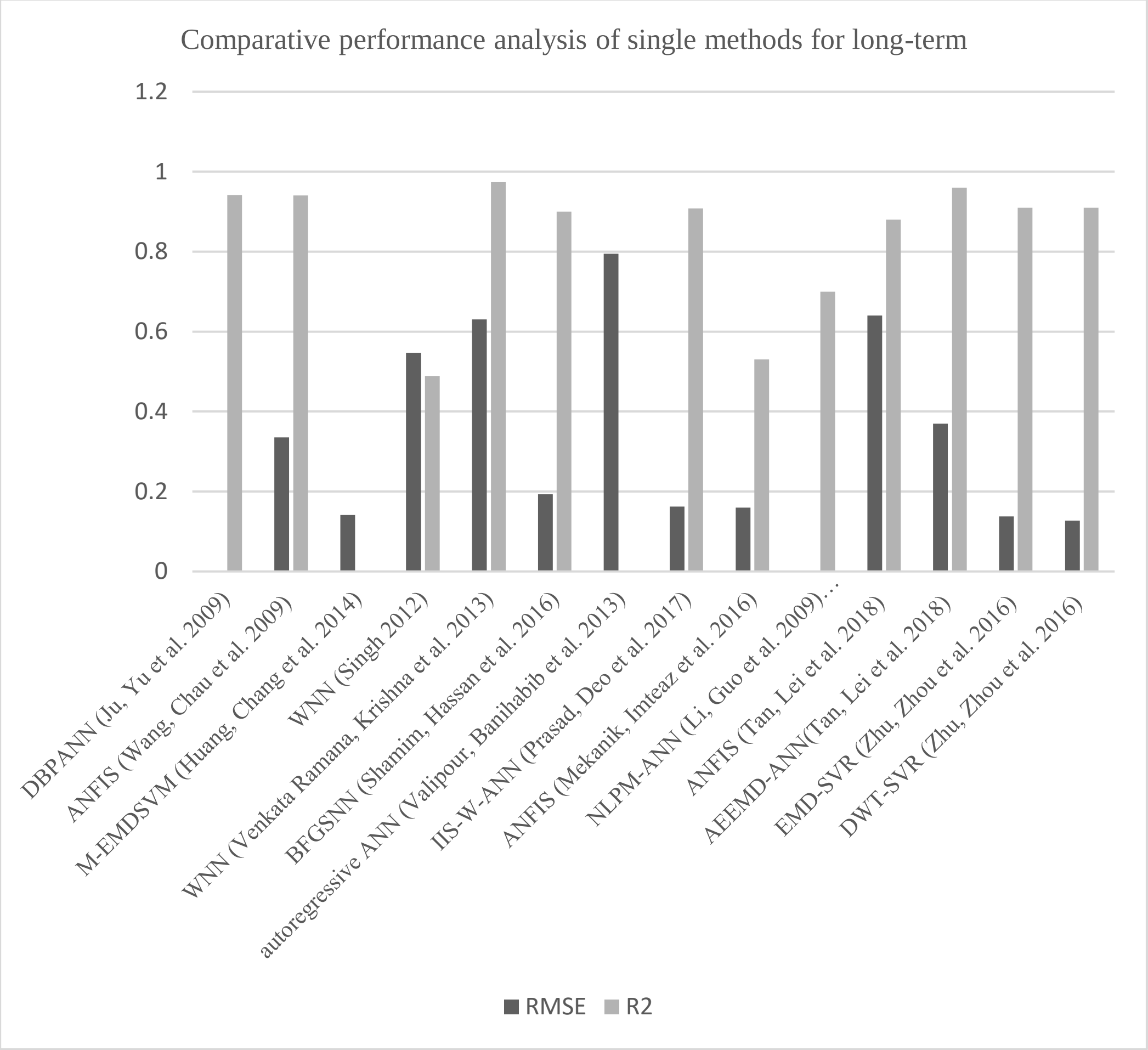

**Figure 10.** Comparative performance analysis of hybrid methods of ML for short-term prediction.

This paper suggests that the drawbacks to major ML methods in terms of accuracy, uncertainty, performance, and robustness were improved through the hybridization of ML methods, as well as using an ensemble variation of the ML method. It is expected that this trend represents the future horizon of flood prediction.

## 5. Conclusions

The current state of ML modeling for flood prediction is quite young and in the early stage of advancement. This paper presents an overview of machine learning models used in flood prediction, and develops a classification scheme to analyze the existing literature. The survey represents the performance analysis and investigation of more than 6000 articles. Among them, we identified 180 original and influential articles where the performance and accuracy of at least two machine learning models were compared. To do so, the prediction models were classified into two categories according to lead time, and further divided into categories of hybrid and single methods. The state of the art of these classes was discussed and analyzed in detail, considering the performance comparison of the methods available in the literature. The performance of the methods was evaluated in terms of $R^2$ and RMSE, in addition to the generalization ability, robustness, computation cost, and speed. Despite the promising results already reported in implementing the most popular machine learning methods, e.g., ANNs, SVM, SVR, ANFIS, WNN, and DTs, there was significant research and experimentation for further improvement and advancement. In this context, there were four major trends reported in the literature for improving the quality of prediction. The first was novel hybridization, either through the integration of two or more machine learning methods or the integration of a machine learning method(s) with more conventional means, and/or soft computing. The second was the use of data decomposition techniques for the purpose of improving the quality of the dataset, which highly contributed in improving the accuracy of prediction. The third was the use of an ensemble of methods, which dramatically increased the generalization ability of the models and decreased the uncertainty of prediction. The fourth was the use of add-on optimizer algorithms to improve the quality of machine learning algorithms, e.g., for better tuning the ANNs to reach optimal neuronal architectures. It is expected that, through these four key technologies, flood prediction will witness significant improvements for both short-term and long-term predictions. Surely, the advancement of these novel ML methods depends highly on the proper usage of soft computing techniques in designing novel learning algorithms. This fact was discussed in the paper, and the soft computing techniques were introduced as the main contributors in developing hybrid ML methods of the future.

Here, it is also worth mentioning that the multidisciplinary nature of this work was the most challenging difficulty to overcome in this paper. Having contributions from the coauthors of both realms of ML and hydrology was the key to success. Furthermore, the novel search methodology and the creative taxonomy and classification of the ML methods led to the original achievement of the paper.

For future work, conducting a survey on spatial flood prediction using machine learning models is highly encouraged. This important aspect of flood prediction was excluded from our paper due to the nature of modeling methodologies and the datasets used in predicting the location of floods. Nevertheless, the recent advancements in machine learning models for spatial flood analysis revolutionized this particular realm of flood forecasting, which requires separate investigation.

**Author Contributions:** P.O., the machine learning expert, contributed to investigation, methodology, supervision, communication, resources, data curation, and project management. A.M. contributed to the original draft preparation, data collection, formal data analysis, investigation, critical review, and final revisions, K.C., the hydrology expert, contributed to supervision, validation, revision, discussion, resources, improvement, and advice.

**Funding:** This research was funded by the Norwegian University of Science and Technology AI Lab and the European Research Consortium for Informatics and Mathematics (ERCIM).

**Acknowledgments:** Dr. Amir Mosavi contributed to this research during his tenure through an ERCIM Alain Bensoussan Fellowship Program under supervision of Prof. Pinar Ozturk.

**Conflicts of Interest:** The authors declare no conflicts of interest.

**Nomenclatures**

| | |
|---|---|
| WMO | World meteorological organization |
| GCM | Global circulation models |
| SPOTA | Seasonal Pacific Ocean temperature analysis |
| ANN | Artificial neural networks |
| POTA | Pacific Ocean temperature analysis |
| QPE | Quantitative precipitation estimation |
| CLIM | Climatology average method |
| EOF | Empirical orthogonal function |
| MLR | Multiple linear regressions |
| QPF | Quantitative precipitation forecasting |
| MNLR | Multiple nonlinear regressions |
| ML | Machine learning |
| MLR | Multiple linear regression |
| ANN | Neural networks |
| WNN | Wavelet-based neural network |
| ARIMA | Auto regressive integrated moving average |
| USGS | United States Geological Survey |
| FFA | Flood frequency analyses |
| QRT | Quantile regression techniques |
| SPOTA | Seasonal Pacific Ocean temperature analysis |
| SVM | Support vector machines |
| LS-SVM | Least-square support vector machines |
| AI | Artificial intelligence |
| VRM | Vector Regression Machine |
| FFNN | Feed-forward neural network |
| FBNN | Feed-backward networks |
| MLP | Multilayer perceptron |
| ANFIS | Adaptive neuro-fuzzy inference system |
| BPNN | Backpropagation neural network |
| SVR | Support vector regression |
| LR | Linear regression |
| FIS | Fuzzy inference system |
| CART | Classification and regression tree |
| LMT | Logistic model trees |
| NWP | Numerical weather prediction |
| NBT | Naive Bayes trees |
| ARMA | Autoregressive moving averaging |
| REPT | Reduced-error pruning trees |
| DT | Decision tree |
| ELM | Extreme learning machine |
| EPS | Ensemble prediction systems |
| SNIP | Source normalized impact per paper |
| SRM | Structural risk minimization |
| AR | Autoregressive |
| SJR | SCImago journal rank |
| ARMAX | Linear autoregressive moving average with exogenous inputs |
| LMT | Logistic model trees |
| ARMA | Autoregressive moving averaging |
| ADT | Alternating decision trees |
| NARX network | Nonlinear autoregressive network with exogenous inputs |
| RMSE | Root-mean-square error |
| RFFA | Regional flood frequency analysis |
| NLR | Nonlinear regression |
| AR | Autoregressive |
| WARM | Wavelet autoregressive model |
| NLR-R | Nonlinear regression with regionalization approach |
| E | Nash Sutcliffe index |
| FR | Frequency ratio |
| SOM | Self-organizing map |
| CHIM | Cluster-based hybrid inundation model |
| FFRM | Flash flood routing model |
| KGE | Kling-Gupta efficiency |
| AME | ANN-based monsoon rainfall enhancement |
| SSNN | State-space neural network |
| SSL | Suspended sediment load |
| NSE | Nash–Sutcliffe efficiency |
| E-CHAID | Exhaustive CHAID |
| CHAID | Chi-squared automatic interaction detector |
| CLIM | Climatology average model |
| HEC–HMS | Hydrologic engineering center–hydrologic modeling system |
| SOM | Self-organizing map |
| PBIAS | Percent bias |
| NLPM | Nonlinear perturbation model |
| RF | Rotation forest |
| KSOFM-NNM | Kohonen self-organizing feature maps neural networks model |
| DBP | Division-based backpropagation |
| DBPANN | DBP neural network |
| NLPM-ANN | Nonlinear perturbation model based on neural network |
| GRNNM | Generalized regression neural networks model |
| IIS | Iterative input selection |
| EEMD | Ensemble empirical mode decomposition |
| ANNE | Artificial neural network ensembles |
| DWT | Discrete wavelet transform |
| SFF | Seasonal flood forecasting |
| MP | Water monitoring points |
| WBANN | Wavelet–bootstrap–ANN |
| HBI | Hilsenhoff's biotic index |
| RT | Regression trees |
| EMD | Empirical mode decomposition |
| LLR | Local linear regression |
| BFGS | Broyden Fletcher Goldfarb Shanno |
| M-EMD | Modified empirical mode decomposition |
| IIS | Iterative input selection |
| SAR | Seasonal first-order autoregressive |
| BFGSNN | Broyden Fletcher Goldfarb Shanno neural network |
| GRNN | Artificial neural networks including generalized regression network |
| T–S | Takagi–Sugeno |
| WLGP | Wavelet linear genetic programming |
| E | Nash coefficients |
| TSC-T–S | Clustering based Takagi–Sugeno |
| TCs | Tropical cyclones |
| PCA | Principal component analysis |

## References

1. Danso-Amoako, E.; Scholz, M.; Kalimeris, N.; Yang, Q.; Shao, J. Predicting dam failure risk for sustainable flood retention basins: A generic case study for the wider greater manchester area. *Comput. Environ. Urban Syst.* **2012**, *36*, 423–433.
2. Xie, K.; Ozbay, K.; Zhu, Y.; Yang, H. Evacuation zone modeling under climate change: A data-driven method. *J. Infrastruct. Syst.* **2017**, *23*, 04017013.
3. Pitt, M. *Learning Lessons from the 2007 Floods*; Cabinet Office: London, UK, 2008.
4. Lohani, A.K.; Goel, N.; Bhatia, K. Improving real time flood forecasting using fuzzy inference system. *J. Hydrol.* **2014**, *509*, 25–41.
5. Mosavi, A.; Bathla, Y.; Varkonyi-Koczy, A. Predicting the Future Using Web Knowledge: State of the Art Survey. In *Recent Advances in Technology Research and Education*; Springer: Cham, Switzerland, 2017; pp. 341–349.
6. Zhao, M.; Hendon, H.H. Representation and prediction of the indian ocean dipole in the poama seasonal forecast model. *Q. J. R. Meteorol. Soc.* **2009**, *135*, 337–352.
7. Borah, D.K. Hydrologic procedures of storm event watershed models: A comprehensive review and comparison. *Hydrol. Process.* **2011**, *25*, 3472–3489.
8. Costabile, P.; Costanzo, C.; Macchione, F. A storm event watershed model for surface runoff based on 2D fully dynamic wave equations. *Hydrol. Process.* **2013**, *27*, 554–569.
9. Cea, L.; Garrido, M.; Puertas, J. Experimental validation of two-dimensional depth-averaged models for forecasting rainfall–runoff from precipitation data in urban areas. *J. Hydrol.* **2010**, *382*, 88–102.

10. Fernández-Pato, J.; Caviedes-Voullième, D.; García-Navarro, P. Rainfall/runoff simulation with 2D full shallow water equations: Sensitivity analysis and calibration of infiltration parameters. *J. Hydrol.* **2016**, *536*, 496–513.
11. Caviedes-Voullième, D.; García-Navarro, P.; Murillo, J. Influence of mesh structure on 2D full shallow water equations and SCS curve number simulation of rainfall/runoff events. *J. Hydrol.* **2012**, *448*, 39–59.
12. Costabile, P.; Costanzo, C.; Macchione, F. Comparative analysis of overland flow models using finite volume schemes. *J. Hydroinform.* **2012**, *14*, 122.
13. Xia, X.; Liang, Q.; Ming, X.; Hou, J. An efficient and stable hydrodynamic model with novel source term discretization schemes for overland flow and flood simulations. *Water Resour. Res.* **2017**, *53*, 3730–3759.
14. Liang, X.; Lettenmaier, D.P.; Wood, E.F.; Burges, S.J. A simple hydrologically based model of land surface water and energy fluxes for general circulation models. *J. Geophys. Res. Atmos.* **1994**, *99*, 14415–14428.
15. Costabile, P.; Macchione, F. Enhancing river model set-up for 2-D dynamic flood modelling. *Environ. Model. Softw.* **2015**, *67*, 89–107.
16. Nayak, P.; Sudheer, K.; Rangan, D.; Ramasastri, K. Short-term flood forecasting with a neurofuzzy model. *Water Resour. Res.* **2005**, *41*, doi:10.1029/2004WR003562.
17. Kim, B.; Sanders, B.F.; Famiglietti, J.S.; Guinot, V. Urban flood modeling with porous shallow-water equations: A case study of model errors in the presence of anisotropic porosity. *J. Hydrol.* **2015**, *523*, 680–692.
18. Van den Honert, R.C.; McAneney, J. The 2011 brisbane floods: Causes, impacts and implications. *Water* **2011**, *3*, 1149–1173.
19. Lee, T.H.; Georgakakos, K.P. Operational rainfall prediction on meso-$\gamma$ scales for hydrologic applications. *Water Resour. Res.* **1996**, *32*, 987–1003.
20. Shrestha, D.; Robertson, D.; Wang, Q.; Pagano, T.; Hapuarachchi, H. Evaluation of numerical weather prediction model precipitation forecasts for short-term streamflow forecasting purpose. *Hydrol. Earth Syst. Sci.* **2013**, *17*, 1913–1931.
21. Bellos, V.; Tsakiris, G. A hybrid method for flood simulation in small catchments combining hydrodynamic and hydrological techniques. *J. Hydrol.* **2016**, *540*, 331–339.
22. Bout, B.; Jetten, V. The validity of flow approximations when simulating catchment-integrated flash floods. *J. Hydrol.* **2018**, *556*, 674–688.

23. Costabile, P.; Macchione, F.; Natale, L.; Petaccia, G. Flood mapping using lidar dem. Limitations of the 1-D modeling highlighted by the 2-D approach. *Nat. Hazards* **2015**, *77*, 181–204.
24. Valipour, M.; Banihabib, M.E.; Behbahani, S.M.R. Parameters estimate of autoregressive moving average and autoregressive integrated moving average models and compare their ability for inflow forecasting. *J. Math. Stat.* **2012**, *8*, 330–338.
25. Adamowski, J.; Fung Chan, H.; Prasher, S.O.; Ozga-Zielinski, B.; Sliusarieva, A. Comparison of multiple linear and nonlinear regression, autoregressive integrated moving average, artificial neural network, and wavelet artificial neural network methods for urban water demand forecasting in montreal, Canada. *Water Resour. Res.* **2012**, *48*, doi:10.1029/2010WR009945.
26. Valipour, M.; Banihabib, M.E.; Behbahani, S.M.R. Comparison of the ARMA, ARIMA, and the autoregressive artificial neural network models in forecasting the monthly inflow of Dez dam reservoir. *J. Hydrol.* **2013**, *476*, 433–441.
27. Chow, V.T.; Maidment, D.R.; Larry, W. *Mays. Applied hydrology*; International Edition; MacGraw-Hill, Inc.: New York, NY, USA, 1988; p. 149.
28. Aziz, K.; Rahman, A.; Fang, G.; Shrestha, S. Application of artificial neural networks in regional flood frequency analysis: A case study for australia. *Stoch. Environ. Res. Risk Assess.* **2014**, *28*, 541–554.
29. Kroll, C.N.; Vogel, R.M. Probability distribution of low streamflow series in the united states. *J. Hydrol. Eng.* **2002**, *7*, 137–146.
30. Mackey, B.P.; Krishnamurti, T. Ensemble forecast of a typhoon flood event. *Weather Forecast.* **2001**, *16*, 399–415.
31. Haddad, K.; Rahman, A. Regional flood frequency analysis in eastern australia: Bayesian GLS regression-based methods within fixed region and ROI framework–quantile regression vs. Parameter regression technique. *J. Hydrol.* **2012**, *430*, 142–161.
32. Thompson, S.A. *Hydrology for Water Management*; CRC Press: Boca Raton, FL, USA, 2017.
33. Kerkhoven, E.; Gan, T.Y. A modified ISBA surface scheme for modeling the hydrology of Athabasca river basin with GCM-scale data. *Adv. Water Resour.* **2006**, *29*, 808–826.
34. Burnash, R.J.; Ferral, R.L.; McGuire, R.A. *A Generalized Streamflow Simulation System, Conceptual Modeling for Digital Computers*; Stanford University: Stanford, CA, USA, 1973.
35. Yamazaki, D.; Kanae, S.; Kim, H.; Oki, T. A physically based description of floodplain inundation dynamics in a global river routing model. *Water Resour. Res.* **2011**, *47*, doi:10.1029/2010WR009726.

36. Fawcett, R.; Stone, R. A comparison of two seasonal rainfall forecasting systems for Australia. *Aust. Meteorol. Oceanogr. J.* **2010**, *60*, 15–24.
37. Mekanik, F.; Imteaz, M.; Gato-Trinidad, S.; Elmahdi, A. Multiple regression and artificial neural network for long-term rainfall forecasting using large scale climate modes. *J. Hydrol.* **2013**, *503*, 11–21.
38. Mosavi, A.; Rabczuk, T.; Varkonyi-Koczy, A. R. Reviewing the novel machine learning tools for materials design. In *Recent Advances in Technology Research and Education*; Springer: Cham, Switzerland, 2017; pp. 50–58.
39. Abbot, J.; Marohasy, J. Input selection and optimisation for monthly rainfall forecasting in Queensland, Australia, using artificial neural networks. *Atmos. Res.* **2014**, *138*, 166–178.
40. Fox, N.I.; Wikle, C.K. A bayesian quantitative precipitation nowcast scheme. *Weather Forecast.* **2005**, *20*, 264–275.
41. Merz, B.; Hall, J.; Disse, M.; Schumann, A. Fluvial flood risk management in a changing world. *Nat. Hazards Earth Syst. Sci.* **2010**, *10*, 509–527.
42. Xu, Z.; Li, J. Short-term inflow forecasting using an artificial neural network model. *Hydrol. Process.* **2002**, *16*, 2423–2439.
43. Ortiz-García, E.; Salcedo-Sanz, S.; Casanova-Mateo, C. Accurate precipitation prediction with support vector classifiers: A study including novel predictive variables and observational data. *Atmos. Res.* **2014**, *139*, 128–136.
44. Kim, S.; Matsumi, Y.; Pan, S.; Mase, H. A real-time forecast model using artificial neural network for after-runner storm surges on the Tottori Coast, Japan. *Ocean Eng.* **2016**, *122*, 44–53.
45. Mosavi, A.; Edalatifar, M. A Hybrid Neuro-Fuzzy Algorithm for Prediction of Reference Evapotranspiration. In *Recent Advances in Technology Research and Education*; Springer: Cham, Switzerland, 2018; pp. 235–243.
46. Dineva, A.; Várkonyi-Kóczy, A.R.; Tar, J.K. Fuzzy expert system for automatic wavelet shrinkage procedure selection for noise suppression. In Proceedings of the 2014 IEEE 18th International Conference on Intelligent Engineering Systems (INES), Tihany, Hungary, 3–5 July 2014; pp. 163–168.
47. Suykens, J.A.; Vandewalle, J. Least squares support vector machine classifiers. *Neural Process. Lett.* **1999**, *9*, 293–300.
48. Gizaw, M.S.; Gan, T.Y. Regional flood frequency analysis using support vector regression under historical and future climate. *J. Hydrol.* **2016**, *538*, 387–398.

49. Taherei Ghazvinei, P.; Hassanpour Darvishi, H.; Mosavi, A.; Yusof, K.B.W.; Alizamir, M.; Shamshirband, S.; Chau, K.W. Sugarcane growth prediction based on meteorological parameters using extreme learning machine and artificial neural network. *Eng. Appl. Comput. Fluid Mech.* **2018**, 12, 738–749.
50. Kasiviswanathan, K.; He, J.; Sudheer, K.; Tay, J.-H. Potential application of wavelet neural network ensemble to forecast streamflow for flood management. *J. Hydrol.* **2016**, *536*, 161–173.
51. Ravansalar, M.; Rajaee, T.; Kisi, O. Wavelet-linear genetic programming: A new approach for modeling monthly streamflow. *J. Hydrol.* **2017**, *549*, 461–475.
52. Mosavi, A.; Rabczuk, T. Learning and intelligent optimization for material design innovation. In *Learning and Intelligent Optimization*; Springer: Cham, Switzerland, 2017; pp. 358–363.
53. Dandagala, S.; Reddy, M.S.; Murthy, D.S.; Nagaraj, G. Artificial neural networks applications in groundwater hydrology—A review. *Artif. Intell. Syst. Mach. Learn.* **2017**, *9*, 182–187.
54. Deka, P.C. Support vector machine applications in the field of hydrology: A review. *Appl. Soft Comput.* **2014**, *19*, 372–386.
55. Fotovatikhah, F.; Herrera, M.; Shamshirband, S.; Chau, K.-W.; Faizollahzadeh Ardabili, S.; Piran, M.J. Survey of computational intelligence as basis to big flood management: Challenges, research directions and future work. *Eng. Appl. Comput. Fluid Mech.* **2018**, *12*, 411–437.
56. Faizollahzadeh Ardabili, S.; Najafi, B.; Alizamir, M.; Mosavi, A.; Shamshirband, S.; Rabczuk, T. Using SVM-RSM and ELM-RSM Approaches for Optimizing the Production Process of Methyl and Ethyl Esters. Energies **2018**, 11, 2889.
57. Tsai, L.T.; Yang, C.-C. Improving measurement invariance assessments in survey research with missing data by novel artificial neural networks. *Expert Syst. Appl.* **2012**, *39*, 10456–10464.
58. Sivapalan, M.; Blöschl, G.; Merz, R.; Gutknecht, D. Linking flood frequency to long-term water balance: Incorporating effects of seasonality. *Water Resour. Res.* **2005**, *41*, doi:10.1029/2004WR003439.
59. Maier, H.R.; Jain, A.; Dandy, G.C.; Sudheer, K.P. Methods used for the development of neural networks for the prediction of water resource variables in river systems: Current status and future directions. *Environ. Model. Softw.* **2010**, *25*, 891–909.
60. Lafdani, E.K.; Nia, A.M.; Pahlavanravi, A.; Ahmadi, A.; Jajarmizadeh, M. Research article daily rainfall-runoff prediction and simulation using ANN, ANFIS and conceptual hydrological MIKE11/NAM models. *Int. J. Eng. Technol.* **2013**, *1*, 32–50.

61. Collier, C. Flash flood forecasting: What are the limits of predictability? *Q. J. R. Meteorol. Soc.* **2007**, *133*, 3–23.
62. Seo, D.-J.; Breidenbach, J. Real-time correction of spatially nonuniform bias in radar rainfall data using rain gauge measurements. *J. Hydrometeorol.* **2002**, *3*, 93–111.
63. Grecu, M.; Krajewski, W. A large-sample investigation of statistical procedures for radar-based short-term quantitative precipitation forecasting. *J. Hydrol.* **2000**, *239*, 69–84.
64. Maddox, R.A.; Zhang, J.; Gourley, J.J.; Howard, K.W. Weather radar coverage over the contiguous united states. *Weather Forecast.* **2002**, *17*, 927–934.
65. Campolo, M.; Andreussi, P.; Soldati, A. River flood forecasting with a neural network model. *Water Resour. Res.* **1999**, *35*, 1191–1197.
66. Prakash, O.; Sudheer, K.; Srinivasan, K. Improved higher lead time river flow forecasts using sequential neural network with error updating. *J. Hydrol. Hydromech.* **2014**, *62*, 60–74.
67. Shu, C.; Ouarda, T. Regional flood frequency analysis at ungauged sites using the adaptive neuro-fuzzy inference system. *J. Hydrol.* **2008**, *349*, 31–43.
68. Ashrafi, M.; Chua, L.H.C.; Quek, C.; Qin, X. A fully-online neuro-fuzzy model for flow forecasting in basins with limited data. *J. Hydrol.* **2017**, *545*, 424–435.
69. Yu, P.-S.; Yang, T.-C.; Chen, S.-Y.; Kuo, C.-M.; Tseng, H.-W. Comparison of random forests and support vector machine for real-time radar-derived rainfall forecasting. *J. Hydrol.* **2017**, *552*, 92–104.
70. Nourani, V.; Baghanam, A.H.; Adamowski, J.; Kisi, O. Applications of hybrid wavelet–artificial intelligence models in hydrology: A review. *J. Hydrol.* **2014**, *514*, 358–377.
71. Zadeh, M.R.; Amin, S.; Khalili, D.; Singh, V.P. Daily outflow prediction by multi layer perceptron with logistic sigmoid and tangent sigmoid activation functions. *Water Resour. Manag.* **2010**, *24*, 2673–2688.
72. Li, L.; Xu, H.; Chen, X.; Simonovic, S. Streamflow forecast and reservoir operation performance assessment under climate change. *Water Resour. Manag.* **2010**, *24*, 83.
73. Wu, C.; Chau, K.-W. Data-driven models for monthly streamflow time series prediction. *Eng. Appl. Artif. Intell.* **2010**, *23*, 1350–1367.
74. Sulaiman, J.; Wahab, S.H. Heavy rainfall forecasting model using artificial neural network for flood prone area. In *It Convergence and Security 2017*; Springer: Singapore, 2018; pp. 68–76.

75. Kar, A.K.; Lohani, A.K.; Goel, N.K.; Roy, G.P. Development of flood forecasting system using statistical and ANN techniques in the downstream catchment of mahanadi basin, india. *J. Water Resour. Prot.* **2010**, *2*, 880.
76. Jain, A.; Prasad Indurthy, S. Closure to "comparative analysis of event-based rainfall-runoff modeling techniques—Deterministic, statistical, and artificial neural networks" by ASHU JAIN and SKV prasad indurthy. *J. Hydrol. Eng.* **2004**, *9*, 551–553.
77. Lohani, A.; Kumar, R.; Singh, R. Hydrological time series modeling: A comparison between adaptive neuro-fuzzy, neural network and autoregressive techniques. *J. Hydrol.* **2012**, *442*, 23–35.
78. Tanty, R.; Desmukh, T.S. Application of artificial neural network in hydrology—A review. *Int. J. Eng. Technol. Res.* **2015**, *4*, 184–188.
79. Kişi, O. Streamflow forecasting using different artificial neural network algorithms. *J. Hydrol. Eng.* **2007**, *12*, 532–539.
80. Shamseldin, A.Y. Artificial neural network model for river flow forecasting in a developing country. *J. Hydroinform.* **2010**, *12*, 22–35.
81. Badrzadeh, H.; Sarukkalige, R.; Jayawardena, A. Impact of multi-resolution analysis of artificial intelligence models inputs on multi-step ahead river flow forecasting. *J. Hydrol.* **2013**, *507*, 75–85.
82. Smith, J.; Eli, R.N. Neural-network models of rainfall-runoff process. *J. Water Resour. Plan. Manag.* **1995**, *121*, 499–508.
83. Taormina, R.; Chau, K.-W.; Sethi, R. Artificial neural network simulation of hourly groundwater levels in a coastal aquifer system of the Venice Lagoon. *Eng. Appl. Artif. Intell.* **2012**, *25*, 1670–1676.
84. Thirumalaiah, K.; Deo, M. River stage forecasting using artificial neural networks. *J. Hydrol. Eng.* **1998**, *3*, 26–32.
85. Panagoulia, D. Artificial neural networks and high and low flows in various climate regimes. *Hydrol. Sci. J.* **2006**, *51*, 563–587.
86. Panagoulia, D.; Tsekouras, G.; Kousiouris, G. A multi-stage methodology for selecting input variables in ann forecasting of river flows. *Glob. Nest J.* **2017**, *19*, 49–57.
87. Deo, R.C.; Şahin, M. Application of the artificial neural network model for prediction of monthly standardized precipitation and evapotranspiration index using hydrometeorological parameters and climate indices in Eastern Australia. *Atmos. Res.* **2015**, *161*, 65–81.

88. Coulibaly, P.; Dibike, Y.B.; Anctil, F. Downscaling precipitation and temperature with temporal neural networks. *J. Hydrometeorol.* **2005**, *6*, 483–496.
89. Schoof, J.T.; Pryor, S. Downscaling temperature and precipitation: A comparison of regression-based methods and artificial neural networks. *Int. J. Climatol.* **2001**, *21*, 773–790.
90. Hassan, Z.; Shamsudin, S.; Harun, S.; Malek, M.A.; Hamidon, N. Suitability of ANN applied as a hydrological model coupled with statistical downscaling model: A case study in the northern area of peninsular Malaysia. *Environ. Earth Sci.* **2015**, *74*, 463–477.
91. Zhang, J.-S.; Xiao, X.-C. Predicting chaotic time series using recurrent neural network. *Chin. Phys. Lett.* **2000**, *17*, 88.
92. Huang, G.-B.; Zhu, Q.-Y.; Siew, C.-K. Extreme learning machine: Theory and applications. *Neurocomputing* **2006**, *70*, 489–501.
93. Lima, A.R.; Cannon, A.J.; Hsieh, W.W. Forecasting daily streamflow using online sequential extreme learning machines. *J. Hydrol.* **2016**, *537*, 431–443.
94. Yaseen, Z.M.; Jaafar, O.; Deo, R.C.; Kisi, O.; Adamowski, J.; Quilty, J.; El-Shafie, A. Stream-flow forecasting using extreme learning machines: A case study in a semi-arid region in Iraq. *J. Hydrol.* **2016**, *542*, 603–614.
95. Sahoo, G.; Ray, C. Flow forecasting for a Hawaii stream using rating curves and neural networks. *J. Hydrol.* **2006**, *317*, 63–80.
96. Kim, S.; Singh, V.P. Flood forecasting using neural computing techniques and conceptual class segregation. *JAWRA J. Am. Water Resour. Assoc.* **2013**, *49*, 1421–1435.
97. Rumelhart, D.E.; Hinton, G.E.; Williams, R.J. Learning representations by back-propagating errors. *Nature* **1986**, *323*, 533.
98. Riad, S.; Mania, J.; Bouchaou, L.; Najjar, Y. Rainfall-runoff model using an artificial neural network approach. *Math. Comput. Model.* **2004**, *40*, 839–846.
99. Senthil Kumar, A.; Sudheer, K.; Jain, S.; Agarwal, P. Rainfall-runoff modelling using artificial neural networks: Comparison of network types. *Hydrol. Process. Int. J.* **2005**, *19*, 1277–1291.
100. Zadeh, L.A. Soft computing and fuzzy logic. In *Fuzzy Sets, Fuzzy Logic, and Fuzzy Systems: Selected Papers by Lotfi a Zadeh*; World Scientific: Singapore, 1996; pp. 796–804.
101. Choubin, B.; Khalighi-Sigaroodi, S.; Malekian, A.; Ahmad, S.; Attarod, P. Drought forecasting in a semi-arid watershed using climate signals: A neuro-fuzzy modeling approach. *J. Mt. Sci.* **2014**, *11*, 1593–1605.

102. Choubin, B.; Khalighi-Sigaroodi, S.; Malekian, A.; Kişi, Ö. Multiple linear regression, multi-layer perceptron network and adaptive neuro-fuzzy inference system for forecasting precipitation based on large-scale climate signals. *Hydrol. Sci. J.* **2016**, *61*, 1001–1009.
103. Bogardi, I.; Duckstein, L. The fuzzy logic paradigm of risk analysis. In *Risk-Based Decisionmaking in Water Resources X*; American Society of Civil Engineers: Reston, VA, USA, 2003; pp. 12–22.
104. See, L.; Openshaw, S. A hybrid multi-model approach to river level forecasting. *Hydrol. Sci. J.* **2000**, *45*, 523–536.
105. Tiwari, M.K.; Chatterjee, C. Development of an accurate and reliable hourly flood forecasting model using wavelet–bootstrap–ANN (WBANN) hybrid approach. *J. Hydrol.* **2010**, *394*, 458–470.
106. Guimarães Santos, C.A.; da Silva, G.B.L. Daily streamflow forecasting using a wavelet transform and artificial neural network hybrid models. *Hydrol. Sci. J.* **2014**, *59*, 312–324.
107. Supratid, S.; Aribarg, T.; Supharatid, S. An integration of stationary wavelet transform and nonlinear autoregressive neural network with exogenous input for baseline and future forecasting of reservoir inflow. *Water Resour. Manag.* **2017**, *31*, 4023–4043.
108. Shoaib, M.; Shamseldin, A.Y.; Melville, B.W. Comparative study of different wavelet based neural network models for rainfall–runoff modeling. *J. Hydrol.* **2014**, *515*, 47–58.
109. Dubossarsky, E.; Friedman, J.H.; Ormerod, J.T.; Wand, M.P. Wavelet-based gradient boosting. *Stat. Comput.* **2016**, *26*, 93–105.
110. Partal, T. Wavelet regression and wavelet neural network models for forecasting monthly streamflow. *J. Water Clim. Chang.* **2017**, *8*, 48–61.
111. Shafaei, M.; Kisi, O. Predicting river daily flow using wavelet-artificial neural networks based on regression analyses in comparison with artificial neural networks and support vector machine models. *Neural Comput. Appl.* **2017**, *28*, 15–28.
112. Kumar, S.; Tiwari, M.K.; Chatterjee, C.; Mishra, A. Reservoir inflow forecasting using ensemble models based on neural networks, wavelet analysis and bootstrap method. *Water Resour. Manag.* **2015**, *29*, 4863–4883.
113. Seo, Y.; Kim, S.; Kisi, O.; Singh, V.P. Daily water level forecasting using wavelet decomposition and artificial intelligence techniques. *J. Hydrol.* **2015**, *520*, 224–243.
114. Sudhishri, S.; Kumar, A.; Singh. J. K. Comparative Evaluation of Neural Network and Regression Based Models to Simulate Runoff and Sediment Yield in an Outer Himalayan Watershed. *Journal of Agricultural Science and Technology*. **2016**, 18, 681–694.

115. Heasrt, M.A.; Dumais, S.T.; Osuna, E.; Platt, J.; Scholkopf, B. Support vector machines. *IEEE Intell. Syst. Their Appl.* **1998**, *13*, 18–28.
116. Vapnik, V.; Mukherjee, S. Support vector method for multivariate density estimation. *Adv. Neural Inf. Process. Syst.* **2000**, 4, 659–665.
117. Li, S.; Ma, K.; Jin, Z.; Zhu, Y. A new flood forecasting model based on SVM and boosting learning algorithms. In Proceedings of the 2016 IEEE Congress on Evolutionary Computation (CEC), Vancouver, BC, Canada, 24–29 July 2016; pp. 1343–1348.
118. Dehghani, M.; Saghafian, B.; Nasiri Saleh, F.; Farokhnia, A.; Noori, R. Uncertainty analysis of streamflow drought forecast using artificial neural networks and Monte-Carlo simulation. *Int. J. Climatol.* **2014**, *34*, 1169–1180.
119. Dibike, Y.B.; Velickov, S.; Solomatine, D.; Abbott, M.B. Model induction with support vector machines: Introduction and applications. *J. Comput. Civ. Eng.* **2001**, *15*, 208–216.
120. Nayak, M.A.; Ghosh, S. Prediction of extreme rainfall event using weather pattern recognition and support vector machine classifier. *Theor. Appl. Climatol.* **2013**, *114*, 583–603.
121. Granata, F.; Gargano, R.; de Marinis, G. Support vector regression for rainfall-runoff modeling in urban drainage: A comparison with the EPA's storm water management model. *Water* **2016**, *8*, 69.
122. Gong, Y.; Zhang, Y.; Lan, S.; Wang, H. A comparative study of artificial neural networks, support vector machines and adaptive neuro fuzzy inference system for forecasting groundwater levels near lake okeechobee, Florida. *Water Resour. Manag.* **2016**, *30*, 375–391.
123. Jajarmizadeh, M.; Lafdani, E.K.; Harun, S.; Ahmadi, A. Application of SVM and swat models for monthly streamflow prediction, a case study in South of Iran. *KSCE J. Civ. Eng.* **2015**, *19*, 345–357.
124. Bao, Y.; Xiong, T.; Hu, Z. Multi-step-ahead time series prediction using multiple-output support vector regression. *Neurocomputing* **2014**, *129*, 482–493.
125. Bray, M.; Han, D. Identification of support vector machines for runoff modelling. *J. Hydroinform.* **2004**, *6*, 265–280.
126. Tehrany, M.S.; Pradhan, B.; Mansor, S.; Ahmad, N. Flood susceptibility assessment using GIS-based support vector machine model with different kernel types. *Catena* **2015**, *125*, 91–101.
127. Kisi, O.; Parmar, K.S. Application of least square support vector machine and multivariate adaptive regression spline models in long term prediction of river water pollution. *J. Hydrol.* **2016**, *534*, 104–112.

128. Liong, S.Y.; Sivapragasam, C. Flood stage forecasting with support vector machines 1. *JAWRA J. Am. Water Resour. Assoc.* **2002**, *38*, 173–186.
129. Sachindra, D.; Huang, F.; Barton, A.; Perera, B. Least square support vector and multi-linear regression for statistically downscaling general circulation model outputs to catchment streamflows. *Int. J. Climatol.* **2013**, *33*, 1087–1106.
130. De'ath, G.; Fabricius, K.E. Classification and regression trees: A powerful yet simple technique for ecological data analysis. *Ecology* **2000**, *81*, 3178–3192.
131. Tehrany, M.S.; Pradhan, B.; Jebur, M.N. Spatial prediction of flood susceptible areas using rule based decision tree (DT) and a novel ensemble bivariate and multivariate statistical models in GIS. *J. Hydrol.* **2013**, *504*, 69–79.
132. Dehghani, M.; Saghafian, B.; Rivaz, F.; Khodadadi, A. Evaluation of dynamic regression and artificial neural networks models for real-time hydrological drought forecasting. *Arabian J. Geosci.* **2017**, *10*, 266.
133. Choubin, B.; Zehtabian, G.; Azareh, A.; Rafiei-Sardooi, E.; Sajedi-Hosseini, F.; Kişi, Ö. Precipitation forecasting using classification and regression trees (CART) model: A comparative study of different approaches. *Environ. Earth Sci.* **2018**, *77*, 314.
134. Choubin, B.; Darabi, H.; Rahmati, O.; Sajedi-Hosseini, F.; Kløve, B. River suspended sediment modelling using the cart model: A comparative study of machine learning techniques. *Sci. Total Environ.* **2018**, *615*, 272–281.
135. Liaw, A.; Wiener, M. Classification and regression by randomforest. *R News* **2002**, *2*, 18–22.
136. Wang, Z.; Lai, C.; Chen, X.; Yang, B.; Zhao, S.; Bai, X. Flood hazard risk assessment model based on random forest. *J. Hydrol.* **2015**, *527*, 1130–1141.
137. Tehrany, M.S.; Pradhan, B.; Jebur, M.N. Flood susceptibility mapping using a novel ensemble weights-of-evidence and support vector machine models in GIS. *J. Hydrol.* **2014**, *512*, 332–343.
138. Bui, D.T.; Tuan, T.A.; Klempe, H.; Pradhan, B.; Revhaug, I. Spatial prediction models for shallow landslide hazards: A comparative assessment of the efficacy of support vector machines, artificial neural networks, kernel logistic regression, and logistic model tree. *Landslides* **2016**, *13*, 361–378.
139. Etemad-Shahidi, A.; Mahjoobi, J. Comparison between m5' model tree and neural networks for prediction of significant wave height in lake superior. *Ocean Eng.* **2009**, *36*, 1175–1181.
140. Dietterich, T.G. Ensemble methods in machine learning. In *International Workshop on Multiple Classifier Systems*; Springer: Berlin/Heidelberg, Germany, 2000; pp. 1–15.

141. Sajedi-Hosseini, F.; Malekian, A.; Choubin, B.; Rahmati, O.; Cipullo, S.; Coulon, F.; Pradhan, B. A novel machine learning-based approach for the risk assessment of nitrate groundwater contamination. *Sci. Total Environ.* **2018**, *644*, 954–962.
142. Moore, K.J.; Kurt, M.; Eriten, M.; McFarland, D.M.; Bergman, L.A.; Vakakis, A.F. Wavelet-bounded empirical mode decomposition for measured time series analysis. *Mech. Syst. Signal Process.* **2018**, *99*, 14–29.
143. Wang, W.-C.; Chau, K.-W.; Xu, D.-M.; Chen, X.-Y. Improving forecasting accuracy of annual runoff time series using ARIMA based on EEMD decomposition. *Water Resour. Manag.* **2015**, *29*, 2655–2675.
144. Al-Musaylh, M.S.; Deo, R.C.; Li, Y.; Adamowski, J.F. Two-phase particle swarm optimized-support vector regression hybrid model integrated with improved empirical mode decomposition with adaptive noise for multiple-horizon electricity demand forecasting. *Appl. Energy* **2018**, *217*, 422–439.
145. Ouyang, Q.; Lu, W.; Xin, X.; Zhang, Y.; Cheng, W.; Yu, T. Monthly rainfall forecasting using EEMD-SVR based on phase-space reconstruction. *Water Resour. Manag.* **2016**, *30*, 2311–2325.
146. Zhang, J.; Hou, G.; Ma, B.; Hua, W. Operating characteristic information extraction of flood discharge structure based on complete ensemble empirical mode decomposition with adaptive noise and permutation entropy. *J. Vib. Control* **2018**, doi:10.1177/1077546317750979.
147. Badrzadeh, H.; Sarukkalige, R.; Jayawardena, A. Hourly runoff forecasting for flood risk management: Application of various computational intelligence models. *J. Hydrol.* **2015**, *529*, 1633–1643.
148. Kim, G.; Barros, A.P. Quantitative flood forecasting using multisensor data and neural networks. *J. Hydrol.* **2001**, *246*, 45–62.
149. Saghafian, B.; Haghnegahdar, A.; Dehghani, M. Effect of ENSO on annual maximum floods and volume over threshold in the southwestern region of Iran. *Hydrol. Sci. J.* **2017**, *62*, 1039–1049.
150. Kourgialas, N.N.; Dokou, Z.; Karatzas, G.P. Statistical analysis and ann modeling for predicting hydrological extremes under climate change scenarios: The example of a small Mediterranean Agro-watershed. *J. Environ. Manag.* **2015**, *154*, 86–101.
151. Panda, R.K.; Pramanik, N.; Bala, B. Simulation of river stage using artificial neural network and mike 11 hydrodynamic model. *Comput. Geosci.* **2010**, *36*, 735–745.

152. Noori, R.; Karbassi, A.; Farokhnia, A.; Dehghani, M. Predicting the longitudinal dispersion coefficient using support vector machine and adaptive neuro-fuzzy inference system techniques. *Environ. Eng. Sci.* **2009**, *26*, 1503–1510.
153. Pereira Filho, A.J.; dos Santos, C.C. Modeling a densely urbanized watershed with an artificial neural network, weather radar and telemetric data. *J. Hydrol.* **2006**, *317*, 31–48.
154. Jingyi, Z.; Hall, M.J. Regional flood frequency analysis for the Gan-Ming River basin in China. *J. Hydrol.* **2004**, *296*, 98–117.
155. Ahmad, S.; Simonovic, S.P. An artificial neural network model for generating hydrograph from hydro-meteorological parameters. *J. Hydrol.* **2005**, *315*, 236–251.
156. Ju, Q.; Yu, Z.; Hao, Z.; Ou, G.; Zhao, J.; Liu, D. Division-based rainfall-runoff simulations with BP neural networks and Xinanjiang model. *Neurocomputing* **2009**, *72*, 2873–2883.
157. Sahoo, G.B.; Ray, C.; De Carlo, E.H. Use of neural network to predict flash flood and attendant water qualities of a mountainous stream on Oahu, Hawaii. *J. Hydrol.* **2006**, *327*, 525–538.
158. Ghose, D.K. *Measuring Discharge Using Back-Propagation Neural Network: A Case Study on Brahmani River Basin*; Springer: Singapore, 2018; pp. 591–598.
159. Pan, H.-X.; Cheng, G.-J.; Cai, L. Comparison of the extreme learning machine with the support vector machine for reservoir permeability prediction. *Comput. Eng. Sci.* **2010**, *2*, 37.
160. Chang, F.-J.; Chen, P.-A.; Lu, Y.-R.; Huang, E.; Chang, K.-Y. Real-time multi-step-ahead water level forecasting by recurrent neural networks for urban flood control. *J. Hydrol.* **2014**, *517*, 836–846.
161. Shen, H.-Y.; Chang, L.-C. Online multistep-ahead inundation depth forecasts by recurrent NARX networks. *Hydrol. Earth Syst. Sci.* **2013**, *17*, 935–945.
162. Bruen, M.; Yang, J. Functional networks in real-time flood forecasting—A novel application. *Adv. Water Resour.* **2005**, *28*, 899–909.
163. Chiang, Y.-M.; Chang, F.-J.; Jou, B.J.-D.; Lin, P.-F. Dynamic ANN for precipitation estimation and forecasting from radar observations. *J. Hydrol.* **2007**, *334*, 250–261.
164. Bhattacharya, B.; Solomatine, D.P. Neural networks and M5 model trees in modelling water level-discharge relationship. *Neurocomputing* **2005**, *63*, 381–396.
165. Heiser, M.; Scheidl, C.; Eisl, J.; Spangl, B.; Hübl, J. Process type identification in torrential catchments in the eastern Alps. *Geomorphology* **2015**, *232*, 239–247.

166. Khosravi, K.; Pham, B.T.; Chapi, K.; Shirzadi, A.; Shahabi, H.; Revhaug, I.; Prakash, I.; Tien Bui, D. A comparative assessment of decision trees algorithms for flash flood susceptibility modeling at haraz watershed, Northern Iran. *Sci. Total Environ.* **2018**, *627*, 744–755.
167. Aichouri, I.; Hani, A.; Bougherira, N.; Djabri, L.; Chaffai, H.; Lallahem, S. River flow model using artificial neural networks. *Energy Procedia* **2015**, *74*, 1007–1014.
168. Torabi, M.; Hashemi, S.; Saybani, M. R.; Shamshirband, S.; Mosavi, A. A Hybrid clustering and classification technique for forecasting short-term energy consumption. *Environ. Prog. Sustain. Energy* **2018**, *47*, doi:10.1002/ep.12934.
169. Adamowski, J.F. Development of a short-term river flood forecasting method for snowmelt driven floods based on wavelet and cross-wavelet analysis. *J. Hydrol.* **2008**, *353*, 247–266.
170. Leahy, P.; Kiely, G.; Corcoran, G. Structural optimisation and input selection of an artificial neural network for river level prediction. *J. Hydrol.* **2008**, *355*, 192–201.
171. Wei, C.C. Soft computing techniques in ensemble precipitation nowcast. *Appl. Soft Comput. J.* **2013**, *13*, 793–805.
172. R Schiffer, R. A.; Rossow, W. B. The International Satellite Cloud Climatology Project (ISCCP): The first project of the world climate research programme. *Bulletin of the American Meteorological Society.* **1983**, 64, 779-784.
173. Castillo, E. Functional networks. *Neural Process. Lett.* **1998**, *7*, 151–159.
174. Jimeno-Sáez, P.; Senent-Aparicio, J.; Pérez-Sánchez, J.; Pulido-Velazquez, D.; María Cecilia, J. Estimation of instantaneous peak flow using machine-learning models and empirical formula in peninsular Spain. *Water* **2017**, *9*, 347.
175. Chang, F.-J.; Chang, Y.-T. Adaptive neuro-fuzzy inference system for prediction of water level in reservoir. *Adv. Water Resour.* **2006**, *29*, 1–10.
176. Mosavi, A.; Lopez, A.; Varkonyi-Koczy, A. R. Industrial Applications of Big Data: State of the Art Survey. In *Recent Advances in Technology Research and Education*; Springer: Cham, Switzerland, 2017; pp. 225–232.
177. Rezaeianzadeh, M.; Tabari, H.; Yazdi, A.A.; Isik, S.; Kalin, L. Flood flow forecasting using ANN, ANFIS and regression models. *Neural Comput. Appl.* **2014**, *25*, 25–37.
178. Tehrany, M.S.; Pradhan, B.; Jebur, M.N. Flood susceptibility analysis and its verification using a novel ensemble support vector machine and frequency ratio method. *Stoch. Environ. Res. Risk Assess.* **2015**, *29*, 1149–1165.

179. Young, C.C.; Liu, W.C.; Wu, M.C. A physically based and machine learning hybrid approach for accurate rainfall-runoff modeling during extreme typhoon events. *Appl. Soft Comput. J.* **2017**, *53*, 205–216.
180. Altunkaynak, A.; Nigussie, T.A. Prediction of daily rainfall by a hybrid wavelet-season-neuro technique. *J. Hydrol.* **2015**, *529*, 287–301.
181. Chang, L.-C.; Shen, H.-Y.; Chang, F.-J. Regional flood inundation nowcast using hybrid SOM and dynamic neural networks. *J. Hydrol.* **2014**, *519*, 476–489.
182. Nanda, T.; Sahoo, B.; Beria, H.; Chatterjee, C. A wavelet-based non-linear autoregressive with exogenous inputs (WNARX) dynamic neural network model for real-time flood forecasting using satellite-based rainfall products. *J. Hydrol.* **2016**, *539*, 57–73.
183. French, J.; Mawdsley, R.; Fujiyama, T.; Achuthan, K. Combining machine learning with computational hydrodynamics for prediction of tidal surge inundation at estuarine ports. *Procedia IUTAM* **2017**, *25*, 28–35.
184. Hong, W.-C. Rainfall forecasting by technological machine learning models. *Appl. Math. Comput.* **2008**, *200*, 41–57.
185. Pan, T.-Y.; Yang, Y.-T.; Kuo, H.-C.; Tan, Y.-C.; Lai, J.-S.; Chang, T.-J.; Lee, C.-S.; Hsu, K.H. Improvement of watershed flood forecasting by typhoon rainfall climate model with an ANN-based southwest monsoon rainfall enhancement. *J. Hydrol.* **2013**, *506*, 90–100.
186. Rajurkar, M.; Kothyari, U.; Chaube, U. Modeling of the daily rainfall-runoff relationship with artificial neural network. *J. Hydrol.* **2004**, *285*, 96–113.
187. Hsu, M.-H.; Lin, S.-H.; Fu, J.-C.; Chung, S.-F.; Chen, A.S. Longitudinal stage profiles forecasting in rivers for flash floods. *J. Hydrol.* **2010**, *388*, 426–437.
188. Pang, B.; Guo, S.; Xiong, L.; Li, C. A nonlinear perturbation model based on artificial neural network. *J. Hydrol.* **2007**, *333*, 504–516.
189. Doycheva, K.; Horn, G.; Koch, C.; Schumann, A.; König, M. Assessment and weighting of meteorological ensemble forecast members based on supervised machine learning with application to runoff simulations and flood warning. *Adv. Eng. Inform.* **2017**, *33*, 427–439.
190. Fleming, S.W.; Bourdin, D.R.; Campbell, D.; Stull, R.B.; Gardner, T. Development and operational testing of a super-ensemble artificial intelligence flood-forecast model for a pacific northwest river. *J. Am. Water Resour. Assoc.* **2015**, *51*, 502–512.

191. Choubin, B.; Khalighi, S.S.; Malekian, A. *Impacts of Large-Scale Climate Signals on Seasonal Rainfall in the Maharlu-Bakhtegan Watershed*; Journal of Range and Watershed Management: Kashan, Iran, 2016.

192. Kisi, O.; Sanikhani, H. Prediction of long-term monthly precipitation using several soft computing methods without climatic data. *Int. J. Climatol.* **2015**, *35*, 4139–4150.

193. Liang, Z.; Li, Y.; Hu, Y.; Li, B.; Wang, J. A data-driven SVR model for long-term runoff prediction and uncertainty analysis based on the Bayesian framework. *Theor. Appl. Climatol.* **2018**, *133*, 137–149.

194. Han, S.; Coulibaly, P. Bayesian flood forecasting methods: A review. *J. Hydrol.* **2017**, *551*, 340–351.

195. Choubin, B.; Moradi, E.; Golshan, M.; Adamowski, J.; Sajedi-Hosseini, F.; Mosavi, A. An Ensemble prediction of flood susceptibility using multivariate discriminant analysis, classification and regression trees, and support vector machines. *Elsevier Sci. Total Environ.* **2018**, *651*, 2087–2096

196. Teng, J.; Jakeman, A.; Vaze, J.; Croke, B.F.; Dutta, D.; Kim, S. Flood inundation modelling: A review of methods, recent advances and uncertainty analysis. *Environ. Model. Softw.* **2017**, *90*, 201–216.

197. Elsafi, S.H. Artificial neural networks (ANNs) for flood forecasting at Dongola station in the river Nile, Sudan. *Alex. Eng. J.* **2014**, *53*, 655–662.

198. Mohammadzadeh, D.; Bazaz, J. B.; Yazd, S. V. J.; Alavi, A. H. Deriving an intelligent model for soil compression index utilizing multi-gene genetic programming. *Springer Environ. Earth Sci.* **2016**, *75*, 262.

199. Singh, P.; Borah, B. Indian summer monsoon rainfall prediction using artificial neural network. *Stoch. Environ. Res. Risk Assess.* **2013**, *27*, 1585–1599.

200. Shamim, M.A.; Hassan, M.; Ahmad, S.; Zeeshan, M. A comparison of artificial neural networks (ANN) and local linear regression (LLR) techniques for predicting monthly reservoir levels. *KSCE J. Civ. Eng.* **2016**, *20*, 971–977.

201. Rezaeian-Zadeh, M.; Tabari, H.; Abghari, H. Prediction of monthly discharge volume by different artificial neural network algorithms in semi-arid regions. *Arabian J. Geosci.* **2013**, *6*, 2529–2537.

202. Gazendam, E.; Gharabaghi, B.; Ackerman, J.D.; Whiteley, H. Integrative neural networks models for stream assessment in restoration projects. *J. Hydrol.* **2016**, *536*, 339–350.

203. Lin, J.-Y.; Cheng, C.-T.; Chau, K.-W. Using support vector machines for long-term discharge prediction. *Hydrol. Sci. J.* **2006**, *51*, 599–612.
204. Cunningham, S.C.; Griffioen, P.; White, M.D.; Nally, R.M. Assessment of ecosystems: A system for rigorous and rapid mapping of floodplain forest condition for Australia's most important river. *Land Degrad. Dev.* **2018**, *29*, 127–137.
205. Ahmad, S.; Kalra, A.; Stephen, H. Estimating soil moisture using remote sensing data: A machine learning approach. *Adv. Water Resour.* **2010**, *33*, 69–80.
206. Prasad, R.D.; Ravinesh, C.; Li, Y.; Maraseni, T. Input selection and performance optimization of ANN-based streamflow forecasts in the drought-prone murray darling basin region using IIS and MODWT algorithm. *Atmos. Res.* **2017**, *197*, 42–63.
207. Cannas, B.; Fanni, A.; Sias, G.; Tronci, S.; Zedda, M.K. River flow forecasting using neural networks and wavelet analysis. *Geophys. Res. Abstr.* **2005**, *7*, 08651.
208. Najafi, B. An Intelligent Artificial Neural Network-Response Surface Methodology Method for Accessing the Optimum Biodiesel and Diesel Fuel Blending Conditions in a Diesel Engine from the Viewpoint of Exergy and Energy Analysis. *Energies* **2018**, *11*, 860.
209. Singh, R.M. Wavelet-ANN model for flood events. In Proceedings of the International Conference on Soft Computing for Problem Solving (SocProS 2011), Patiala, India, 20–22 December 2011; pp. 165–175.
210. Ramana, R.V.; Krishna, B.; Kumar, S.R.; Pandey, N.G. Monthly rainfall prediction using wavelet neural network analysis. *Water Resour. Manag.* **2013**, *27*, 3697–3711.
211. Tantanee, S.; Patamatammakul, S.; Oki, T.; Sriboonlue, V.; Prempree, T. Coupled wavelet-autoregressive model for annual rainfall prediction. *J. Environ. Hydrol.* **2005**, *13*, 124–146.
212. Mekanik, F.; Imteaz, M.A.; Talei, A. Seasonal rainfall forecasting by adaptive network-based fuzzy inference system (ANFIS) using large scale climate signals. *Clim. Dyn.* **2016**, *46*, 3097–3111.
213. Wang, W.C.; Chau, K.W.; Cheng, C.T.; Qiu, L. A comparison of performance of several artificial intelligence methods for forecasting monthly discharge time series. *J. Hydrol.* **2009**, *374*, 294–306.
214. Kisi, O.; Nia, A.M.; Gosheh, M.G.; Tajabadi, M.R.J.; Ahmadi, A. Intermittent streamflow forecasting by using several data driven techniques. *Water Resour. Manag.* **2012**, *26*, 457–474.
215. Li, C.; Guo, S.; Zhang, J. Modified NLPM-ANN model and its application. *J. Hydrol.* **2009**, *378*, 137–141.

216. Huang, S.; Chang, J.; Huang, Q.; Chen, Y. Monthly streamflow prediction using modified EMD-based support vector machine. *J. Hydrol.* **2014**, *511*, 764–775.
217. Zhu, S.; Zhou, J.; Ye, L.; Meng, C. Streamflow estimation by support vector machine coupled with different methods of time series decomposition in the upper reaches of Yangtze River, China. *Environ. Earth Sci.* **2016**, *75*, 531.
218. Bass, B.; Bedient, P. Surrogate modeling of joint flood risk across coastal watersheds. *J. Hydrol.* **2018**, *558*, 159–173.
219. Araghinejad, S.; Azmi, M.; Kholghi, M. Application of artificial neural network ensembles in probabilistic hydrological forecasting. *J. Hydrol.* **2011**, *407*, 94–104.
220. Tan, Q.-F.; Lei, X.-H.; Wang, X.; Wang, H.; Wen, X.; Ji, Y.; Kang, A.-Q. An adaptive middle and long-term runoff forecast model using EEMD-ANN hybrid approach. *J. Hydrol.* **2018**, doi:10.1016/j.jhydrol.2018.01.015.
221. Nosratabadi, S., Mosavi, A., Shamshirband, S., Kazimieras Zavadskas, E., Rakotonirainy, A. and Chau, K.W., 2019. Sustainable business models: A review. Sustainability, 11(6), p.1663.
222. Høverstad, B.A.; Tidemann, A.; Langseth, H.; Öztürk, P. Short-term load forecasting with seasonal decomposition using evolution for parameter tuning. *IEEE Trans. Smart Grid* **2015**, *6*, 1904–1913.
223. Tantithamthavorn, C.; McIntosh, S.; Hassan, A.E.; Matsumoto, K. Automated parameter optimization of classification techniques for defect prediction models. In Proceedings of the 2016 IEEE/ACM 38th International Conference on Software Engineering (ICSE), Austin, TX, USA, 14–22 May 2016; pp. 321–332.
224. Varkonyi-Koczy, A.R. Review on the usage of the multiobjective optimization package of modefrontier in the energy sector. In *Recent Advances in Technology Research and Education*; Springer: Cham, Switzerland, 2017; p. 217.
225. Dineva, A.; Várkonyi-Kóczy, A.R.; Tar, J.K. Anytime fuzzy supervisory system for signal auto-healing. In *Advanced Materials Research*; Trans Tech Publications: Tihany, Hungary, 2015; pp. 269–272.
226. Torabi, M.; Mosavi, A.; Ozturk, P.; Varkonyi-Koczy, A.; Istvan, V. A hybrid machine learning approach for daily prediction of solar radiation. In *Recent Advances in Technology Research and Education*; Springer: Cham, Switzerland, 2018; pp. 266–274.

227. Solgi, A.; Nourani, V.; Pourhaghi, A. Forecasting daily precipitation using hybrid model of wavelet-artificial neural network and comparison with adaptive neurofuzzy inference system (case study: Verayneh station, Nahavand). *Adv. Civ. Eng.* **2014**, *2014*, 279368.
228. Badrzadeh, H.; Sarukkalige, R.; Jayawardena, A. Improving ann-based short-term and long-term seasonal river flow forecasting with signal processing techniques. *River Res. Appl.* **2016**, *32*, 245–256.